\documentclass[10pt,twocolumn,letterpaper]{article}

\usepackage{cvpr}
\usepackage{times}
\usepackage{epsfig}
\usepackage{graphicx}
\usepackage{amsmath}
\usepackage{amsthm}

\def\R{{\rm I} \! {\rm R}}

\newcommand{\SKIP}[1]{} 

\newcommand{\mbegin} {\left [ \begin{array}}

\newcommand{\mend}   {\end{array} \right ]}

\newcommand{\detbegin} {\left | \begin{array}}
\newcommand{\detend}   {\end{array} \right |}

\newcommand{\vbegin} {\left ( \begin{array}{c}}

\newcommand{\vend} {\end{array}\right )}

\def\squareforqed{\hbox{\rlap{$\sqcap$}$\sqcup$}}

\def\qed{\ifmmode\squareforqed\else{\unskip\nobreak\hfil
	\penalty50\hskip1em\null\nobreak\hfil\squareforqed
	\parfillskip=0pt\finalhyphendemerits=0\endgraf}\fi}

\newcommand{\showeqnlabel}{
	\hbox to 0pt{\quad\quad\relax\fbox{\scriptsize\rm\eqnlblx}%
	\gdef\eqnlblx{xxxx}}} \newcommand{\eqnlblx}{}

\def\@eqnnum{\rm (\theequation)\showeqnlabel}

\newcommand{\nofig}[1]{\centerline{\bf Figure here}}

\usepackage{amssymb}
\usepackage{enumerate}
\usepackage{multirow}
\usepackage{threeparttable}
\usepackage[normalem]{ulem}
\usepackage{mathtools}
\usepackage{adjustbox}
\usepackage{tabularx}
\usepackage[linesnumbered,ruled]{algorithm2e}

\newcolumntype{C}[1]{>{\centering\arraybackslash}p{#1}}
\newcommand{\vB}{\mathbf{B}}

\newcommand{\vk}{\mathbf{k}}
\newcommand{\vL}{\mathbf{L}}
\newcommand{\vE}{\mathbf{E}}
\newcommand{\vy}{\mathbf{y}}

\newcommand{\vx}{\mathbf{x}}
\newcommand{\vX}{\Omega}
\newcommand{\vu}{\mathbf{u}}

\newcommand{\vp}{\mathbf{p}}
\newcommand{\vq}{\mathbf{q}}
\newcommand{\prox}{\mathcal{P}}
\newcommand{\rmT}{\mathrm{T}}
\newcommand{\abs}{\mathrm{abs}}

\newcommand{\R}{{\rm I} \! {\rm R}}

\usepackage[pagebackref=true,breaklinks=true,letterpaper=true,colorlinks
,bookmarks=false]{hyperref}

\cvprfinalcopy 

\def\cvprPaperID{1071} 

\ifcvprfinal\fi
\begin{document}

\title{Single Image Optical Flow Estimation with an Event Camera}

\author{Liyuan Pan$^{1,2}$,~Miaomiao Liu$^{1,2}$,~and Richard Hartley$^{1,2}$\\ 
$^{1}$ Australian National University, Canberra, Australia, ~ ~
$^{2}$ Australian Centre for Robotic Vision \\
\tt\small{liyuan.pan}@anu.edu.au
}

\maketitle

\begin{abstract}

Event cameras are bio-inspired sensors that asynchronously report intensity
changes in microsecond resolution. DAVIS can capture high
dynamics of a scene and simultaneously output high temporal resolution events
and low frame-rate intensity images. 
In this paper, we propose a single image (potentially blurred) and events based
optical flow estimation approach. First, we demonstrate how events can be used
to improve flow estimates. To this end, we encode the relation between flow and
events effectively by presenting an event-based photometric consistency
formulation. Then, we consider the special case of image blur caused by high
dynamics in the visual environments and show that including the blur formation
in our model further constrains flow estimation. This is in sharp contrast to
existing works that ignore the blurred images while our 
formulation can naturally handle either blurred or sharp images to achieve
accurate flow estimation. 
Finally, we reduce flow estimation, as well as image deblurring, to an
alternative optimization problem of an objective function using the primal-dual
algorithm. Experimental results on both synthetic and real data (with blurred
and non-blurred images) show the superiority of our model in comparison to
state-of-the-art approaches.

\end{abstract}


\section{Introduction} 
\begin{figure}[t]
\begin{center}
\begin{tabular}{cc}
\includegraphics[width=0.2135\textwidth]{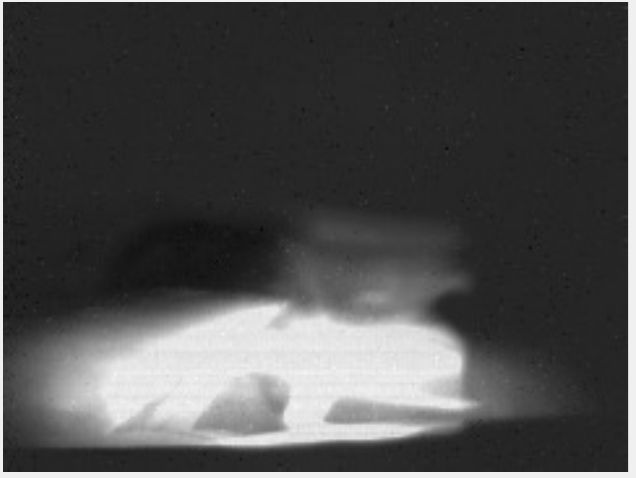}
&\includegraphics[width=0.2135\textwidth]{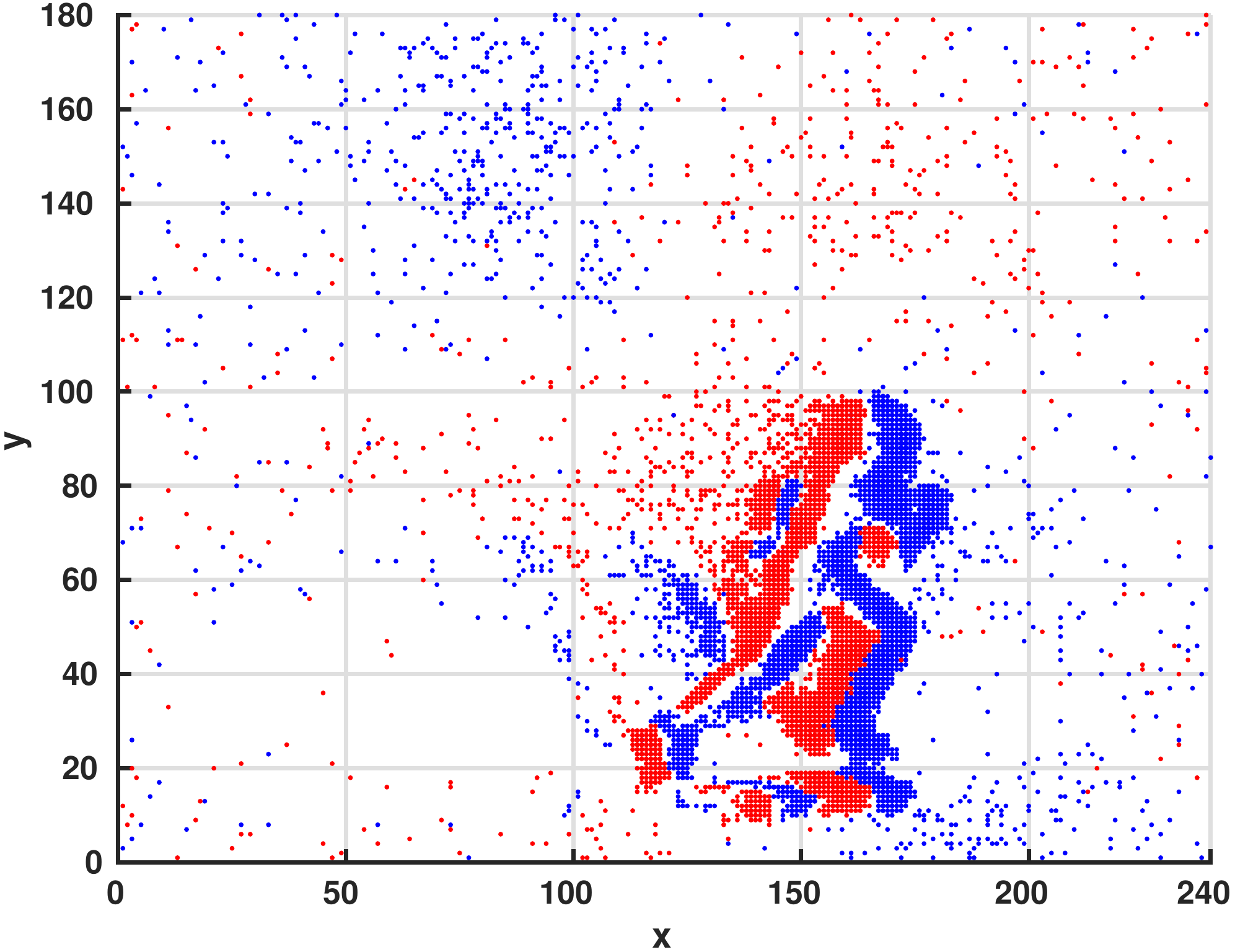}\\
(a) Input image
&(b) Input events\\
\includegraphics[width=0.2135\textwidth]{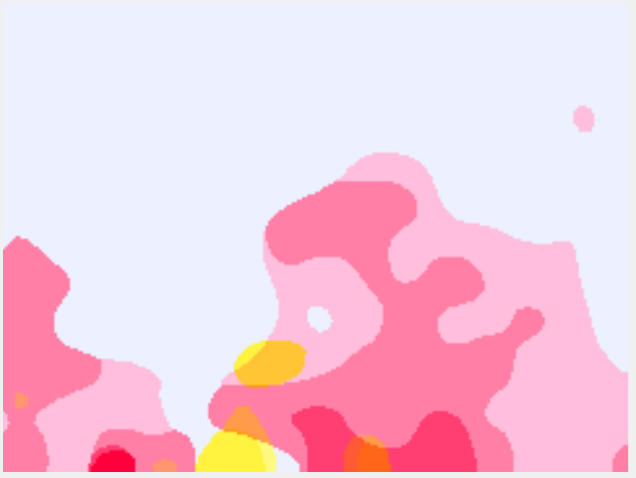}
&\includegraphics[width=0.2135\textwidth]{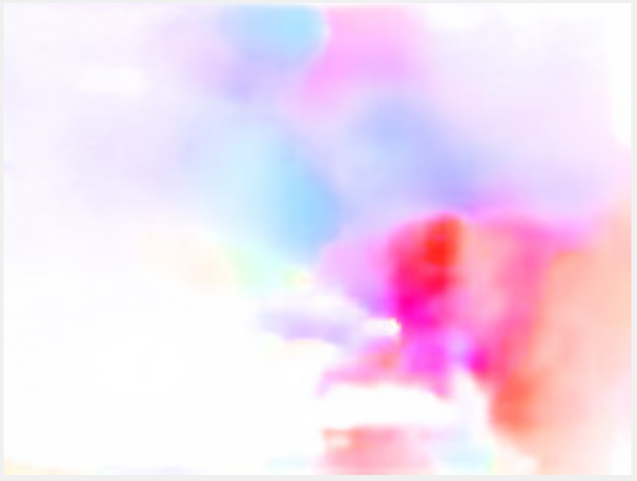}\\
(c) Gong~\etal~\cite{gong2017motion} 
&(d) EV-FlowNet~\cite{Zhu-RSS-18} \\
\includegraphics[width=0.2135\textwidth]{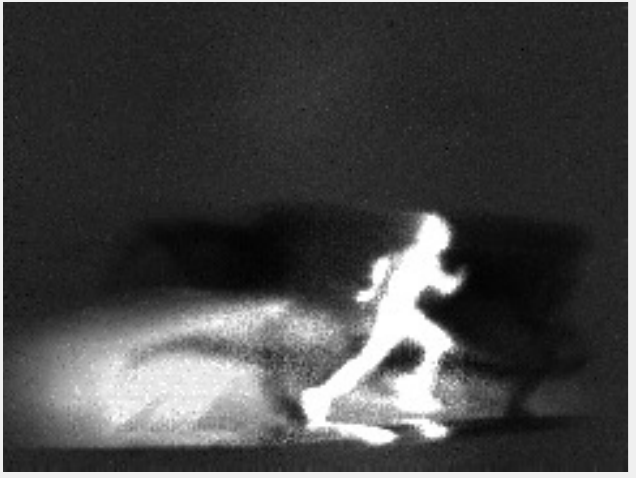}
&\includegraphics[width=0.2135\textwidth]{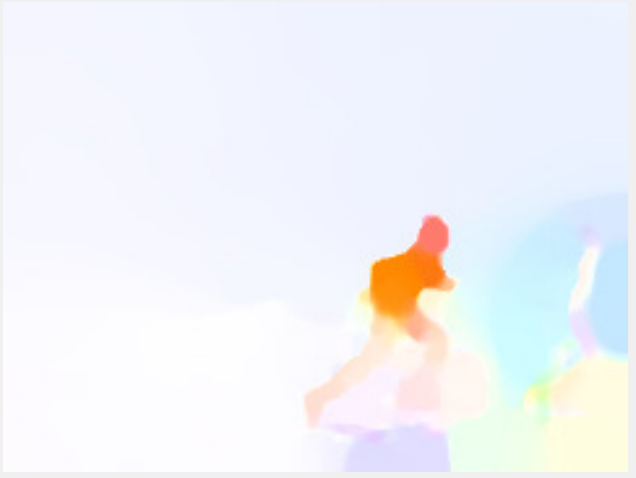}\\
(e) Our deblurred image 
&(f) Our optical flow\\
\end{tabular}
\end{center}
\vspace*{-2 mm}
\caption{\label{fig:nightrun} \em {\bf Optical flow estimation.}
(a) and (b) are the input to our method, where
(a) shows the intensity image from DAVIS, and (b) visualises the
integrated events over a temporal window (blue: positive event; red: negative
event). 
(c) Flow result of \cite{gong2017motion} by using a single blurred image. 
(d) Flow result of \cite{Zhu-RSS-18}, by using events.
(e) and (f) are our results. Our methods is able to handle large motion scenery. 
(Best viewed on screen).
}
\end{figure}

Event cameras (such as DVS \cite{lichtsteiner2008128} and DAVIS
\cite{brandli2014240}) measure intensity changes at each pixel independently
with microsecond accuracy. 
Unlike conventional cameras recording images at a fixed frame rate, event
cameras trigger the event whenever the change in intensity at a given pixel exceeds
a preset threshold.
Event cameras are gaining attention for their high temporal resolution,
robustness to low lighting and highly dynamic scenes which can be used for
tasks such as tracking \cite{rebecq2016evo,Gehrig18eccv}, deblurring
\cite{Pan_2019_CVPR}, and SLAM 
\cite{Kim16eccv,kueng2016low,vidal2018ultimate}.
However, standard vision algorithms cannot be applied to event cameras directly.
Hence, new
methods are required to be tailored to event cameras and unlock their
potential.
In this paper, we aim to show how events can improve flow estimates,
even with a blurred image.

Optical flow estimation is an active topic in the computer vision community and
serves as the backbone for event-based moving object
segmentation~\cite{stoffregen2019event}, human pose
estimation~\cite{calabrese2019dhp19}, and action
recognition~\cite{Amir_2017_CVPR}. 
Traditional flow estimation approaches~\cite{horn1981determining,jason2016back,Yin_2018_CVPR} 
are proposed based on the brightness consistency
assumption for corresponding pixels across the image pair, and cannot handle the
asynchronous event data \cite{gallego2019event}. 
A common trend~\cite{Bardow16cvpr, Gallego_2018_CVPR, zhu2019unsupervised,gehrig2019end} 
to estimate flow is from events only. 
However, events are sparse spatially, flow computed at regions with no events
are less reliable than those computed at regions with events (\ie ,at edges)~\cite{liu2018abmof}. 
Hence, several methods tends to fuse the intensity information and events~\cite{Bardow16cvpr,barranco2014contour} to estimate flow. 

To this end, we aim to utilize the output of DAVIS, which is events and
intensity images, to improve optical flow estimates. 
A straightforward idea is to reconstruct images from events~\cite{Pan_2019_CVPR,Rebecq_2019_CVPR}, 
and then compute flow directly from the reconstructed image. 
While the generated flow is noisy inherently, it shows the potential to estimate
flow by using the image and its event streams (seeing Fig. \ref{fig:alex}). 
Unfortunately, this approach neglects the inherent connection between flow and
events. 
Thus, we introduce an event-based photometric consistency in our model to encode
the relation between flow and event data. 
Different from Zhu \etal~\cite{Zhu-RSS-18} that exploit images as the
supervision signal for a self-supervised learning framework only, we fully
explore the relation between events and flow to formulate our model. 

On the other hand, while intensity images are effective for flow estimation,
output images of event cameras tend to contain blur artefacts due to dynamic
visual environment. It makes flow estimation even more challenging as
brightness constancy may not hold for blurred images (seeing
Fig.~\ref{fig:nightrun}). 
Unlike existing methods, we explore the relationship between flow and blurred
image formation which provides more constraints to flow estimation. 
In a nutshell, our model shows the potential of event cameras for single image
flow estimation, and can also work under blurred condition by joint sharp image
and optical flow estimation.

\vspace*{0.5mm}
In summary, our main contributions are
\vspace*{-1.0mm}

\begin{enumerate}[1)]

\vspace*{-2.2mm}
\item We propose a method for optical flow estimation from a single image
(blurred potentially) and its event data for the event camera (DAVIS).

\vspace*{-2.2mm}
\item We introduce a~\emph{event-based brightness constancy} constraint on absolute intensity to
encode the relation between optical flow and the event data. Besides, we utilize
the blur formation model in our objective function to handle optical flow
estimation on the blurred image.

\vspace*{-2.2mm}
\item Experimental results in both real and synthetic datasets show our method
can successfully handle complex real-world flow estimation, depicting
fast-moving objects, camera motions, and uncontrolled lighting conditions.
\end{enumerate}

\section{Related Work}

In this section, we review works for flow estimation from event cameras, images,
and event-based image reconstruction which could be used for flow estimation. We
further discuss a few works for image deblurring related to flow.%

\vspace*{0.5mm}
{\noindent{\bf Event camera based flow estimation.}} 
Benosman \etal \cite{benosman2012asynchronous} propose an adaptation of the
gradient-based Lucas-Kanade algorithm based on DVS. In \cite{benosman2013event},
they assume that the flow orientation and amplitude can be estimated using a
local differential approach on the surface defined by coactive events. 
They work well for sharp edges and monochromatic blocks but fail with dense
textures, thin lines, and more complicated scenes. 
Barranco \etal \cite{barranco2015bio} propose a more expensive phase-based
method for high-frequency texture regions and trying to reconstruct the
intensity signals to avoid the problem with textured edges. 
Bardow \etal \cite{Bardow16cvpr} jointly reconstruct intensity image and
estimate flow based on events by minimizing their objective function. However,
accuracy relies on the quality of the reconstructed image. 
Gallego \etal \cite{Gallego_2018_CVPR} present a unifying framework to estimate
flow by finding the point trajectories on each image plane that are best aligned
with events. 
Zhu \etal propose EV-FlowNet~\cite{Zhu-RSS-18}, an event-based
flow estimation approach using a self-supervised deep learning pipeline. The
event data are represented as 2D frames to feed the network. While images from the
sensor are used as a supervision signal, the blur effect is ignored which is shown
to be useful for flow estimation in our framework.
In \cite{zhu2019unsupervised}, they further use another event format to train
two networks to predict flow, camera ego-motion, and depth for static scenery.
Then, they use predictions to remove motion blur from event streams which shows
the potential of blurring to improve the flow estimate accuracy. However, flow
computed at those constant brightness regions is still less reliable. 

\vspace*{0.5mm}
{\noindent{\bf Image-based flow estimation.}}
One promising direction is to learn optical flow with CNNs
\cite{dosovitskiy2015flownet,jason2016back,Yin_2018_CVPR} by video. 
FlowNet 2.0~\cite{Ilg_2017_CVPR} develops a stacked architecture that includes
warping of the second image with the intermediate flow.
PWC-Net~\cite{Sun_2018_CVPR} uses the current flow estimate to warp the CNN
features of the second image. It then uses the warped features and features of
the first image to construct a cost volume to estimate flow.
SelFlow~\cite{Liu_2019_CVPR} is based on distilling reliable flow estimations
from non-occluded pixels, and using these predictions to guide optical flow
learning for hallucinated occlusions. 
Several deep learning-driven works attempt to use a single image to estimate
flow~\cite{walker2015dense, rosello2016predicting, endo2019animating}. Walker
\etal \cite{walker2015dense} use CNN to predict dense flow, while they assume
the image is static.

\vspace*{0.5mm}
{\noindent{\bf Event-based image reconstruction.}} 
Image reconstruction \cite{Rebecq_2019_CVPR, Wang_2019_CVPR,pan2019bringing} from events can be
treated as the data preparation step for traditional image-based flow estimation
methods. However, this ignores that the event can contribute to flow estimation.
To reconstruct the image with more details, several methods attempt to combine
events with intensity images 
\cite{Brandli14iscas, Scheerlinck18accv,Pan_2019_CVPR}. 
Pan~\etal~\cite{Pan_2019_CVPR} propose an {{Event-based Double Integral (EDI)}}
model to fuse an image with its events to reconstruct a high frame rate video.
In our paper, we combine the EDI model and state-of-the-art optical flow
estimation methods to serve as baselines of our approach.

\vspace*{0.5mm}
{\noindent{\bf Image deblurring.}}
As the flow accuracy highly depends on the quality of the image, a better-restored
image also relies on the quality of the estimated flow. 
Researchers attempt to use flow to estimate the spatial-varying blur kernel and
then restore images
~\cite{xu2015blind, kim2014segmentation,hyun2015generalized,sellent2016stereo,Pan_2017_CVPR,pan2019joint,pan2019single}. 
Recently, learning-based methods have brought significant improvements in
image deblurring 
\cite{gong2017motion, Nah2019RNN, Zhou2019Stereodeblur}. Gong
\etal~\cite{gong2017motion} directly estimate flow from a blurred image by a
fully-convolutional neural network (FCN) and recover the sharp image from the
estimated flow. 
It is still a challenging problem for
dynamic scene deblurring. Our estimated flow from a single
image and events are more robust and the model generalizes well to handle blurred images
from complex scenery.

\section{Variational Approach}
We start with reviewing variational approaches for optical flow estimation from
a pair of images. Define as $\vu=(u,v)$ to be an optical flow field, and
$\vu(\vx) = (u_{\vx}, v_{\vx})^{\rmT}$ its value at a given pixel $\vx$. From a
reference time $f$ to $t$, the brightness constancy can be written as
\vspace*{-1mm}
%
\begin{equation}  \label{eq:opc}
\vL(\vx,f) = \vL(\vx+\vu(\vx),t) \ ,
\end{equation}
%
where $\vu \in \R^{H \times W \times 2}$, and $\vL \in \R^{H\times W}$ is the
latent image. Here, $H,\ W$ are the image size. Let the intensity of pixel $\vx
= (x,y)^{\rmT}$ at time $f$ be denoted by $\vL(\vx,f)$. 
As equation~\eqref{eq:opc} is under-determined, regularization terms are
introduced to solve optical flow. Horn and Schunck \cite{horn1981determining}
studied a variational formulation of the problem,
\vspace*{-1mm}
%
\begin{equation}\label{eq:horn}
\min_{\vu} \int_{\vX} \|\nabla\vu(\vx)\|^2 \, d\vx 
+\int_{\vX} (\vL(\vx,f)-\vL(\vx+\vu(\vx),t))^2 \,d\vx \ ,
\end{equation}
%
where $\|\cdot\|$ is the standard $l^2$ norm, $\vX$ denotes the image
domain, and $\nabla \vu \in \R^{H \times W \times 4}$. The first term penalizes
high variations in $\vu$ to obtain smooth optical flow fields. The second term
enforces the brightness constancy constraint (BCC). Here, we denote $\nabla \vu(\vx)$
as 
%
\vspace*{-1mm}
%
\begin{equation}\nonumber
\begin{split}
\nabla \vu(\vx)&=\left(\frac{\partial u(\vx)}{\partial x} ,\ 
\frac{\partial u(\vx)}{\partial y} ,\ 
\frac{\partial v(\vx)}{\partial x} ,\ 
\frac{\partial v(\vx)}{\partial y} \right)^{\rmT}\ ,\\
\end{split}
\end{equation}
%
where we denote $\nabla \vu (\vx)=( u_{\vx}^{(x)},
u_{\vx}^{(y)},v_{\vx}^{(x)},v_{\vx}^{(y)})^{\rmT}$ for short. Note that (here
and elsewhere) superscripts in brackets represent differentiation with respect
to $x$ or $y$.

\section{Event-based approach}

We aim to estimate flow from a set of events (from time $f$ to
$t$) and a single corresponding gray-scale image (blurred potentially) taken by DAVIS. 
It is noteworthy that flow is defined as a continuously varying motion
field at a flexible time slice of event data, which is different from the
traditional flow defined based on the image frame rate.

To compute flow from events, a potential solution is to estimate flow from the
reconstructed images based on event cameras \cite{Pan_2019_CVPR}. However, it
ignores that events can contribute to flow estimation. In contrast, we observe that 
events provide correspondences of pixels across time, which implicitly defines
flows for pixels with events. It suggests that we should model events directly
in our flow estimation framework. Meanwhile, the intensity image is another
output of DAVIS. However, it is likely blurred due to high dynamics in the
scene. As shown in~\cite{gong2017motion}, the blur artifacts in the image provides
useful information for flow estimation. 

We therefore propose to jointly estimate flow ${\vu}$ and the latent image
${\vL}$ by enforcing 
the brightness constancy by events and the blurred image formation model. In
particular, our energy minimization model is formulated as:
\begin{equation}\label{eq:energyfunc}
     \min_{\vL,\vu} \mu_1 \phi_\textrm{eve}(\vL,\vu) +\mu_2
\phi_\textrm{blur}(\vL,\vu) +  \phi_\textrm{flow}(\nabla\vu) +
\phi_\textrm{im}(\nabla\vL) \ ,
 \end{equation}
where $\mu_1$ and $\mu_2$ are weight parameters, $\phi_\textrm{eve}$ enforces
the BCC by event, $\phi_\textrm{blur}$
enforces the blurred image formation process, $\phi_\textrm{flow}$ and
$\phi_\textrm{im}$ enforces the smoothness of the estimated flow and latent image.
In following sections, we include details for the objective function in Eq.~\eqref{eq:energyfunc}.

\subsection{Brightness Constancy by Event Data $\phi_\textrm{eve}$}

In case of the output data from DAVIS, we represent Eq.~\eqref{eq:opc} in a
different way. 
Besides images, each {\em event} is denoted by $(\vx,t,\sigma)$. 
Polarity $\sigma = \pm 1 $ denotes the direction of the intensity change. An
event is fired when a change in the log intensity exceeds a threshold $c$.
%
\vspace*{-1mm}
%
\begin{align} 
\left|\log ( {\vL(\vx,t)})-\log ( {\vL(\vx,t_{ref})} \right|\geq c \ .
\end{align}
%
Here, $t$ is the current timestamp and $t_{ref}$ is the timestamp
of the previous event. When an event is triggered, $t_{ref}$ and
$\vL(\vx,t_{ref})$ at that pixel is updated to a new timestamp and a new
intensity level.
Following the EDI model \cite{Pan_2019_CVPR}, we represent the neighbouring
image as
\vspace*{-1mm}
%
\begin{equation}\label{eq:Ltsigma}
\begin{split}
\vL(\vx,t)\ 
& = \vL(\vx,f) \, \exp( c\,  \vE(\vx,t))\ , \\
\end{split}
\end{equation}
%
where $\vE(\vx,t)$ is the integration of events between time $f$ and $t$ at a
given pixel $\vx$, and we dub $\vE(t)$ as the event frame.

Assume the motion between $\triangle t =t-f$ is small. We adopt a first-order
Taylor expansion to the right-hand side of Eq.~\eqref{eq:opc} and obtain its
approximation%
\vspace*{-1mm}
%
\begin{equation} 
\begin{aligned}
\vL(\vx & +\vu(\vx),f+\triangle t) \\
& \approx \vL(\vx,f) + u_{\vx}\vL(\vx,f)^{(x)} +
v_{\vx}\vL(\vx,f)^{(y)}+\triangle t \ \vL(\vx,f)^{(t)}\\
&= u_{\vx}\vL(\vx,f)^{(x)} + v_{\vx}\vL(\vx,f)^{(y)} +\vL(\vx,t) \ . \\
\end{aligned}
\end{equation}
Back to the left-hand side of Eq.~\eqref{eq:opc}, we have 
\vspace*{-1mm}
%
\begin{equation} 
\begin{split}
\vL(\vx,f)\approx u_{\vx}\vL(\vx,f)^{(x)} + v_{\vx}\vL(\vx,f)^{(y)} +
\vL(\vx,t)\ .
\end{split}
\end{equation}
%
With the event model in Eq.~\eqref{eq:Ltsigma}, we can form the latent image as,
\vspace*{-1mm}
%
\begin{equation}\nonumber
\begin{split}
\vL(\vx,f) 
\approx\ & u_{\vx}\vL(\vx,f)^{(x)} + v_{\vx}\vL(\vx,f)^{(y)}\\ &+
\vL(\vx,f)\exp\left(c\, \vE(\vx,t)\right).
\end{split}
\end{equation}
%
Let {\small $\nabla \vL(\vx,f)=(\vL(\vx,f)^{(x)},\vL(\vx,f)^{(y)})^{\rmT}$}, we
therefore write the event-based photometric constancy constraint as
\vspace*{-1mm}
%
\begin{eqnarray}\label{eq:energyeve}
\boxed{
\begin{aligned}
\phi_\textrm{eve}(\vL,\vu)
=\sum_{\vx \in \vX}\| & \vL(\vx,f)(\exp( c\, \vE(\vx,t))-1)\\
&+[u_{\vx},v_{\vx}]^{\rmT}\nabla {\vL}(\vx,f) \|_1 \ .
\end{aligned}
}
\end{eqnarray}
%
Different with ~\cite{Bardow16cvpr,gehrig2019end,bryner2019event} defining the brightness constancy constraint in the log space, we encode the relation between optical flow and events by our event-based
brightness constancy constraint in terms of the  original absolute intensity space.

\subsection{Blur Image Formation Constraint $\phi_\textrm{blur}$}

\begin{figure}
\begin{center}
\begin{tabular}{cc}
 \hspace{-0.25 cm}
\includegraphics[width=0.217\textwidth]{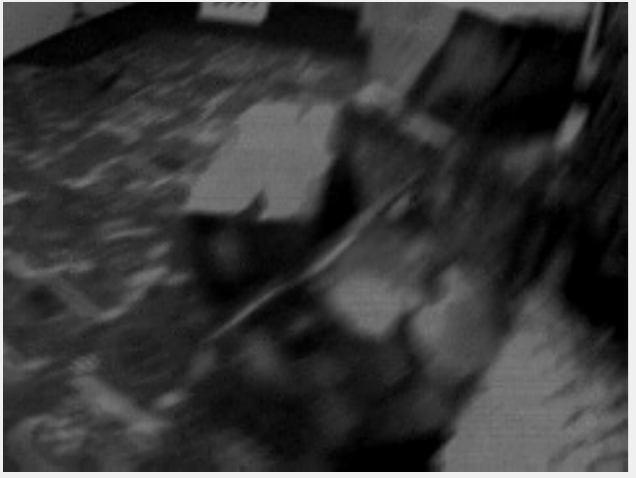}
& \includegraphics[width=0.217\textwidth]{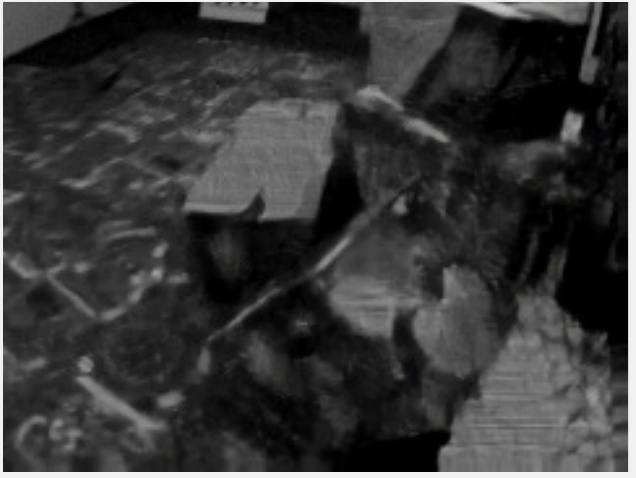}\\
 \hspace{-0.25 cm}
(a) Blurred image
&(b) Deblurred image \cite{Zhang_2019_CVPR}\\
 \hspace{-0.25 cm}
\includegraphics[width=0.217\textwidth]{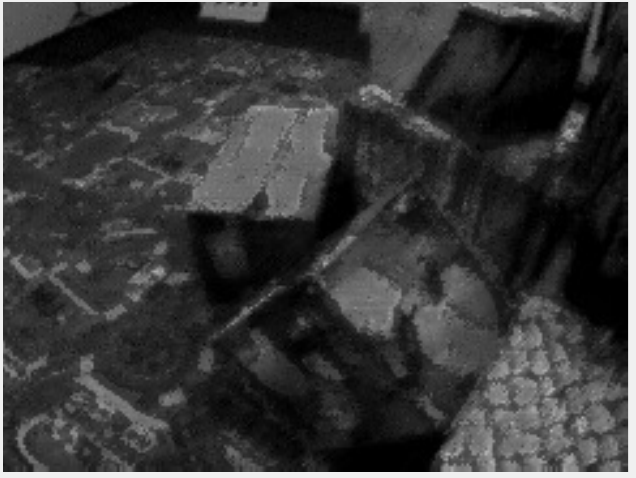}
&\includegraphics[width=0.217\textwidth]{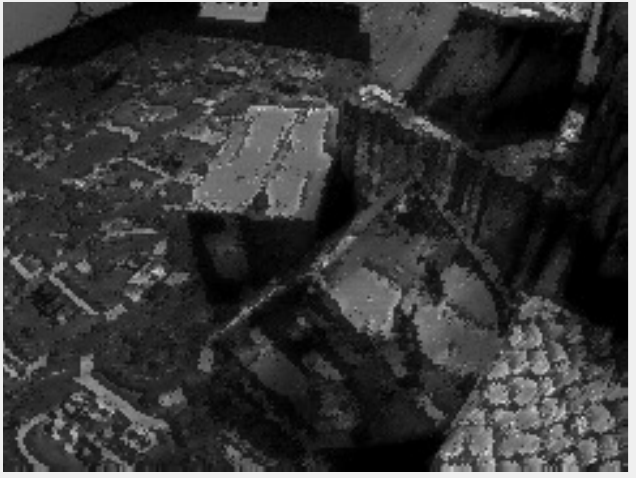} \\
 \hspace{-0.25 cm}
(c) Deblurred image \cite{Pan_2019_CVPR}
&(d) Ours\\
\end{tabular}
\end{center}
\vspace*{-2 mm}
\caption{\label{fig:eth} \em
An example of our deblurring result on the real dataset
\cite{mueggler2017event}. (a) The blurred image. (b) Deblurred by Zhang \etal
\cite{Zhang_2019_CVPR}. (c) Deblurred by EDI \cite{Pan_2019_CVPR}. (d) Ours.
(Best viewed on screen).}
\label{fig:manually}
\end{figure}

In addition to event streams, DAVIS can provide intensity images at a
much lower temporal rate than events.
Images may suffer from motion blur due to the relative motion between the camera
and objects. A general model of blur image formation is given by
\vspace*{-1mm}
%
\begin{eqnarray} \label{eq:bm}
\vB =  \vk \otimes \vL(f) \ ,
\end{eqnarray}
%
where $\vB \in \R ^{H\times W}$ is the blurred image, $\otimes$ is the
convolution operator, and $\vk$ denotes the blur kernel. 
For a dynamic scenario,  
the spatially variant blur kernel is, in principle, defined for each pixel. Then
\vspace*{-1mm}
%
\begin{equation} \label{eq:pixel-bm0}
\begin{aligned}
\vB(\vx) & = \vk(\vx)\otimes \vL(\vx)\ .\\
\end{aligned}
\end{equation}
%
We omit $f$ in the following sections. The convolution of the two matrices is
defined as, 
\vspace*{-1mm}
%
\begin{equation} \label{eq:pixel-bm}
\begin{aligned}
\vB(\vx) &= \sum_{\vy \in \vX} \vk(\vy) \vL(\vx-\vy)\\
&= \sum_{\vy \in \vX} \vk_{\vu'(\vx)}(\vy) \vL(\vx-\vy) \ ,
\end{aligned}
\end{equation}
%
where $\vx,\vy \in \vX$, and $\vk_{\vu'(\vx)} \in \R ^{H\times W}$ is the kernel
map for each pixel. We use the subscript ${\vu'(\vx)}$ to denote the index of
$\vk$ for pixel $\vx$, and 
$\vk_{\vu'(\vx)}(\vy)$ is expressed as
\vspace*{-1mm}
\begin{equation}
\begin{aligned} \label{eq:kimblurKernel}
 k_{\vu'(\vx)}&(\vy)
=&\left\{
\begin{gathered}
\begin{aligned}
&\frac{1}{|\vu'(\vx)|},
&\;{\rm if}\;\vy=\alpha\vu'(\vx),\ |\alpha|\leq\frac{1}{2}\\
& {\bf 0}, & {\rm otherwise}\ ,
\end{aligned}
\end{gathered}\right.
\end{aligned}
\end{equation}
%
where {\small ${\vu'(\vx)=\lambda \vu(\vx)}$} denotes flow during the exposure
time $T$, and {\small $\lambda = T/\triangle t$}. It follows our assumption that
flow during a small time interval has a constant velocity. Furthermore, each
element of the kernel is non-negative and the sum of it is equal to one. Note
that the kernel defined in Eq.~\eqref{eq:kimblurKernel} allows us to handle blurred images with a long exposure time $T$, as well as sharp images with
short exposure time. When $T$ is small, $\theta$ is small enough to result in a Dirac delta function as a blur kernel (\eg convolving a signal
with the delta function leaves the signal unchanged). 
The blur image formation constraint is denoted as
\vspace*{-1mm}
%
\begin{equation}\label{eq:phicolor}
\boxed{
\phi_\textrm{blur}(\vL,\vu) =\sum_{\vx,\vy \in \vX} \|\vk_{\vu'(\vx)}(\vy)
\vL(\vx-\vy)-\vB(\vx)\|^2 \ ,
}
\end{equation}
%
which can handle the blurred and sharp image in a unified framework.

\subsection{Smoothness Term $\phi_\textrm{flow}$, and $\phi_\textrm{im}$}
In general, conventional flow estimation models assume that flow vectors vary
smoothly and have sparse discontinuities at edges of the
image~\cite{hyun2013optical}. Smoothness terms aim to regularize flow and the image
by minimizing the difference between neighbouring pixels. 
For any pixel $\vx$, vector {\small $w(\vx)=(w_{\vx}^{x},w_{\vx}^{y})\in\R^2$},
and {\small $\nabla \vu(\vx)\in\R^4$}, define
\vspace*{-1mm}
%
\begin{equation}\nonumber
\begin{aligned}
w(\vx)\nabla \vu(\vx)
= \left(w_{\vx}^{x} u_{\vx}^{(x)} , w_{\vx}^{y} u_{\vx}^{(y)} , w_{\vx}^{x}
v_{\vx}^{(x)} , w_{\vx}^{y} v_{\vx}^{(y)} \right)^{\rmT} \ .
\end{aligned}
\end{equation}
%
Putting all the pixels together, we define $w\nabla\vu$, where {\small  $w\in\R^{H
\times W \times 2 }$} and {\small $\nabla \vu\in\R^{H \times W \times 4 }$}. 

Our flow cost is defined as
\vspace*{-1mm}
%
\begin{equation}
\boxed{
\begin{aligned}
\phi_\textrm{flow}(\nabla \vu) 
=\|w\nabla \vu\|_{1,2} 
= \sum_{\vx \in \vX}\|w(\vx)\nabla \vu(\vx)\|\ ,
\end{aligned}
}
\end{equation}
%
which is a mixed $1$-$2$ norm (sum of $2$-norms). We choose weight $w$ where
\vspace*{-1mm}
%
\begin{equation}
\begin{aligned}
w^x = \mu_3 \exp(- (\hat{\vL}^{(x)}/\mu_4)^2 ) \ ,
\end{aligned}
\end{equation}
%
and similarly {\small $w^y$}, constants {\small $\mu_3$} and {\small $\mu_4$}
are weight parameters, and {\small $\hat{\vL}$} is the input image of our
optimization framework. 
In addition, we define an image smoothness term as
\vspace*{-1mm}
%
\begin{equation}
\boxed{
\phi_\textrm{im}(\nabla \vL) = \sum_{\vx \in \vX}\|\nabla\vL(\vx)\|_1 \ .
}
\end{equation}
%

\section{Optimization}
Clearly, Eq.~\eqref{eq:energyfunc} is non-convex with respect to ${\vu}$, and ${\vL}$.
Therefore, we perform the optimization over one variable at a time and optimize
all parameters in an alternating manner.  

\begin{itemize}

\item Fix latent image ${\vL}$, and compute optical flow by optimizing
Eq.~\eqref{eq:sceneFlowEnergy} (See Section~\ref{sec:sceneflow}).

\vspace*{-1.7 mm}
\item Fix optical flow $\vu$, and compute the latent image by optimizing
Eq.~\eqref{eq:latentImageEnergy} (See Section~\ref{sec:deblurring}).
\end{itemize}

Here, we use the primal-dual algorithm 
\cite{pock2009algorithm,pock2009convex,chambolle2011first} 
for its optimal convergence. In the
following section, we describe details for each optimization step. 

\subsection{Optical Flow Estimation} \label{sec:sceneflow}

We fix the image, namely $\vL = \hat{\vL}$, and Eq.~\eqref{eq:energyfunc} reduces to
\begin{equation}\label{eq:sceneFlowEnergy}
    \min_{\vu} \underbrace{\mu_1 \phi_\textrm{eve}(\vu) +\mu_2
\phi_\textrm{blur}(\vu)}_{G(\vu)} +  \underbrace{\phi_\textrm{flow}(\nabla
\vu)}_{F(K\vu)}\ ,
\end{equation}
where $\phi_\textrm{eve}(\vu)$ and $\phi_\textrm{flow}(\nabla\vu)$ are convex,
while $\phi_\textrm{blur}(\vu)$ is non-convex. 
As shown, we separate Eq.~\eqref{eq:sceneFlowEnergy} into $G$ and $F$, 
where $K\vu=w\nabla\vu$ is a linear function and
$F(K\vu)=\|K\vu\|_{1,2}=\phi_\textrm{flow}(\nabla\vu)$. Let $\vu\in X=\R^{2N}$,
and $\nabla\vu \in Y=\R^{4N}$, so $G:X \rightarrow \R$, and $F:Y \rightarrow
\R$, where $N=HW$ is the number of pixels. 
In follows, we treat $\vu$, $\nabla \vu$ as vectors. 
The basis of the primal-dual formulation is to replace $F$ in
Eq.~\eqref{eq:sceneFlowEnergy} by its double Fenchel dual $F^{**}$, so it becomes
$\mathop{\min}_{\vu \in X}(G(\vu)+F^{**}(K\vu))$, which is
\vspace*{-1mm}
%
\begin{equation}\label{eq:saddle}
\begin{split}
&\min_{\vu\in X}\left( G(\vu) + \max_{\vp\in Y} \langle K \vu \ ,\vp \rangle_X
-F^{*}(\vp) \ \right) \ . \\
\end{split}
\end{equation}
%
Recall that the Fenchel dual (convex conjugate) $F^*$ of function $F$ is defined
as
\vspace*{-1mm}
%
\begin{equation}
\begin{split}
F^{*}(\vq) &= \sup_{\vp\in Y} \left(\langle \vp,\vq \rangle-F(\vp)\right)\ ,\\
\end{split}
\end{equation}
%
and that $F=F^{**}$ if $F$ is a convex function (a norm is convex). The
primal-dual algorithm of \cite{chambolle2011first} consists of
iterations starting from initial estimates $\vu^0$, $\vp^0$ and
$\bar{\vu}^0=\vu^0$:
\vspace*{-1mm}
%
\begin{equation}\label{eq:fencheliter}
\begin{split}
\vp^{n+1}&=\prox_{F^*}(\vp^n+\sigma K\bar{\vu}^n)\\
\vu^{n+1}&=\prox_{G}(\vu^n-\tau K^*\vp^{n+1})\\
\bar{\vu}^{n+1}&={\vu}^{n+1}+\theta({\vu}^{n+1}-{\vu}^{n}) \ .
\end{split}
\end{equation}
%
Here $\sigma$ and $\tau$ are weight
parameters, and $\prox(\cdot)$ is the proximal operator
\[
\prox_{g}(x) = \arg \min_{y}(2 g(y)+\|y-x\|^2)~.
\]
The hyperparameter $\theta$ is a number that controls the degree of
`extrapolation'. We use $\theta=1$. We now discuss each step of this algorithm
in the present case.

\vspace*{+1 mm}
\noindent{\bf{Updating $\vp$.}}
It is well known that the Fenchel dual of a norm is the indicator function of
the unit ball in the dual norm. In this case, $F^*(\cdot)$ is a mixed norm
$\|\cdot\|_{1,2}$, and its dual is a norm $\|\cdot\|_{\infty,2}$ (details can be
found in the supplementary material). The indicator function is therefore a
product $B^{N}$ of $N$ Euclidean $2$-balls (each in $\R^4$). More precisely
\vspace*{-1mm}
%
\begin{equation}
F^{*}(\vp) = 
\begin{cases}
0, & \textrm{if} \ \|\vp_{\vx}\|\leq 1 \ \textrm{for \ all}\ \vx\\
+\infty, &\textrm{otherwise}\ .
\end{cases}
\end{equation}
%
The proximal operator $\prox_{F^*}$ is therefore given by
\vspace*{-1mm}
%
\begin{equation}
\begin{split}
F^{*}(\bar{\vp}) &=\arg\min_{\vp\in Y}\left(
2F^*(\vp)+\|\bar{\vp}-\vp\|^2\right)\\
&=\arg \min_{\vp\in B^N}\|\bar{\vp}-\vp\|^2 \ .
\end{split}
\end{equation}
%
In other words, each $\bar{\vp}_{\vx}$ is projected to the nearest point in the
unit ball, given by $\bar{\vp}_{\vx}/(\max(1,\|\bar{\vp}_{\vx}\|))$.

\vspace*{+1 mm}
\noindent{\bf{Updating $\vu$.}}
The update equation from Eq.~\eqref{eq:fencheliter} is 
\vspace*{-1mm}
\begin{equation}\nonumber
\begin{split}
\bar{\vu} &= \vu^n-\tau K^* \vp^{n+1}\\
\vu^{n+1} &= \prox_{\tau G}(\bar{\vu})=\mathop{\arg}\mathop{
\min}_{\vu}\left({2\tau}G(\vu)+ {\|\vu-\bar{\vu}\|^2}\right)\ .
\end{split}
\end{equation}
%
(Note we use $\prox_{\tau G}$ instead of $\prox_G$).
Minimizing by taking derivatives gives {\small $\vu = \bar{\vu}-\tau\nabla
G(\vu)$}. We make the simplifying assumption that $G$ is locally approximated
to first order, and so {\small $\nabla G(\vu)=\nabla G(\vu^n)$}, which leads to
the update step
%
\vspace*{-1mm}
%
\begin{align}
\begin{split}
\vu^{n+1} &= 
\vu^n - \tau\big(\nabla G(\vu^n) + K^*\vp^{n+1}\big) \ ,
\end{split}
\end{align}
%
which is simply gradient descent of Eq.~\eqref{eq:saddle}, fixing $\vp=\vp^{n+1}$.
We obtain Algorithm~\ref{Algorithm:PD}.

%
\begin{algorithm} \label{Algorithm:PD}
{\small 
    \SetKwInOut{Initialization}{Initialization}
    \SetKwInOut{Iterations}{Iterations}
\caption{Primal-Dual Minimization - Flow}
\Initialization{Choose $\tau,\, \sigma >0$, $n=0$, and set
$\bar{\vu}^0=\vu^0$.}
\Iterations{Update $\vu^n$, $\vp^n$, $\bar{\vu}^n$ as follows}

\While {$n < 20$}{

\vspace*{+1 mm}
Dual ascent in $\vp$

$\bar{\vp} = {\vp}^{n} + \sigma K \bar{\vu}^n$, 
$\ \vp_{\vx}^{n+1}={\bar{\vp}_{\vx}}/{\max(1,\|\bar{\vp}_{\vx}\|)} ~\forall \vx$

\vspace*{+1 mm}
Primal descent in $\vu$

$\vu^{n+1} = \vu^n - \tau\big(G(\vu^n) + K^*\vp^{n+1}\big)$


\vspace*{+1 mm}
Extrapolation step

$\bar{\vu}^{n+1} = \vu^{n+1}+(\vu^{n+1}-\vu^{n})$

$n=n+1$
}}
\end{algorithm}

\subsection{Deblurring}\label{sec:deblurring}
We fix optical flow, namely $\vu = \hat{\vu}$, and Eq.~\eqref{eq:energyfunc} reduces
to 
\begin{equation}\label{eq:latentImageEnergy}
     \min_{\vL} \underbrace{\phi_\textrm{im}(\nabla
\vL)}_{F_1(\nabla\vL)}+\underbrace{\mu_1\phi_\textrm{eve}(\vL)}_{F_2(K
\vL)}+\underbrace{\mu_2 \phi_\textrm{blur}(\vL)}_{G(\vL)}\ .
\end{equation}
The convex conjugate $F^*$ is defined as,
\vspace*{-1mm}
\begin{equation}
\begin{split}
F^*(\vp,\vq) &= F_1^*(\vp)+F_2^*(\vq)\ ,
\end{split}
\end{equation}
%
where $\vp\in \R ^{2N}$, and $\vq \in \R ^{N}$. Here, $\nabla \vL \in \R
^{2N}$. The primal-dual update process is expressed as follows,
\vspace*{-1mm}
%
\begin{equation}
\begin{aligned}
\vp^{n+1} &= \frac{\vp^{n}+\gamma \nabla 
\bar{\vL}^n}{\max(1,\abs(\vp^{n}+\gamma \nabla  \bar{\vL}^n))},\\
\vq^{n+1} &= \frac{\vq^{n}+\gamma (\theta_2\bar{\vL}^n + [u,\
v]^{\rmT} \nabla \bar{\vL}^n)}{\max(1,\abs (\vq^{n}+\gamma (\theta_2\bar{\vL}^n +
[u,\ v]^{\rmT} \nabla \bar{\vL}^n) ) )}\ ,\\
\end{aligned}
\end{equation}
%
where $\eta$, $\gamma$ are weight factors, and  $\theta_2=\exp(c\vE(t))-1$. 
\vspace*{-1mm}
%
\begin{equation}
\begin{aligned}
\vL^{n+1} &= \prox_{\eta G}(\bar{\vL}) 
=\arg \min_{\vL}\left({2\eta}G(\vL) + {\|\vL-\bar{\vL}\|^2} \right)\ ,\\
\end{aligned}
\end{equation}
%
where $\bar{\vL} = \vL^{n}-\eta (\nabla^* \vp^{n+1} + K^* \vq^{n+1})$.
We obtain Algorithm~\ref{Algorithm:PDD} for the minimization of the proposed
energy function \eqref{eq:latentImageEnergy}.

\begin{algorithm}
\label{Algorithm:PDD}
{\small
    \SetKwInOut{Initialization}{Initialization}
    \SetKwInOut{Iterations}{Iterations}
\caption{Primal-Dual Minimization - Deblurring}
\Initialization{Choose $\gamma,\, \eta >0$, $n=0$, and set
$\bar{\vL}^0=\vL^0$.}
\Iterations{Update $\vL^n$, $\vp^n$, $\vq^n$ as follows}

\While {$ n < 5 $}{

\vspace*{+1 mm}\hspace{-0.2cm}
Dual ascent in $\vp$, $\vq$

\hspace{-0.2cm}
$\bar{\vp} = \vp^{n}+\gamma \nabla \bar{\vL}^n$, 
$\ \bar{\vq} = \vq^{n}+ \gamma (\theta_2 \bar{\vL}^n + [u,\ v]^{\rmT} \nabla
\bar{\vL}^n )$

\hspace{-0.2cm}
$\vp_{\vx}^{n+1}={\bar{\vp}_{\vx}}/{\max(1,\abs(\bar{\vp}_{\vx}))}$~$\forall \vx$

\hspace{-0.2cm}
$\vq_{\vx}^{n+1}={\bar{\vq}_{\vx}}/{\max(1,\abs(\bar{\vq}_{\vx}) )}$~$\forall \vx$

\vspace*{+1 mm}\hspace{-0.2cm}
Primal descent in $\vL$

\hspace{-0.2cm}
$\bar{\vL} = \vL^{n}-\eta (\nabla^* \vp^{n+1} + K^* \vq^{n+1})$, $\
\vL^{n+1} = \prox_{\eta G}(\bar{\vL})$

\vspace*{+1 mm}
\hspace{-0.2cm}
Extrapolation step

\hspace{-0.2cm}
$\bar{\vL}^{n+1} = \vL^{n+1}+(\vL^{n+1}-\vL^{n})$

\hspace{-0.2cm}
$n=n+1$
}}
\vspace*{-1mm}
\end{algorithm}

\vspace*{-3mm}
\section{Experiments}
\begin{table*}
\caption{\label{tab:MVSEC} 
\em Results on the MVSEC \cite{Zhu-RSS-18} and Sintel dataset
\cite{Butler:ECCV:2012}. We evaluate optical flow by Mean Square Error (MSE),
Average Endpoint Error (AEE) and Flow Error metric (FE). The first column `GT
images' means we use two ground-truth images to estimate flow. `EDI image' means
we use two reconstruct images to estimate flow by EDI model. EV-FlowNet
\cite{Zhu-RSS-18} provides a pre-trained model with cropped images ($256\times
256$) and events. Thus, we only show their results that comparing with the cropped ground-truth
flow. Our model achieves competitive results compared with
state-of-the-art methods. Our `AEE' and `FE' metric dropped two times as much as
others.  
}
\centering
\begin{adjustbox}{width=0.96\textwidth}
\begin{tabular}{C{1.5cm}|c|c|c|c|c|c|C{2cm}}
\hline
\multicolumn{8}{c}{MVSEC dataset \cite{Zhu-RSS-18}} \\ \hline
Input          & \multicolumn{2}{c|}{GT images} & \multicolumn{2}{c|}{EDI images and events} &
\multicolumn{2}{c|}{Events} \\ \hline
     & SelFlow~\cite{Liu_2019_CVPR}      &  PWC-Net~\cite{Sun_2018_CVPR}   
 & SelFlow~\cite{Liu_2019_CVPR}      & PWC-Net~\cite{Sun_2018_CVPR}       &
EV-FlowNet~\cite{Zhu-RSS-18}    & Zhu \etal~\cite{zhu2019unsupervised}       &
Ours           \\ \hline
AEE     & 0.5365          & 0.4392          & 1.4232              & 1.3677      
    & 1.3112        & {\bf 0.6975 }       & { 0.9296}          \\ \hline       
MSE     & 0.3708         & 0.1989          & 1.7882              &  1.6135      
   & 1.3501          &  -     & {\bf0.8700}          \\ \hline
FE (\%) & 0.5163          & 0.0938          & 2.5079               & 2.4927     
    & 1.1038         &  1.7500     & {\bf0.4768}          \\ \hline
\multicolumn{8}{c}{Sintel dataset \cite{Butler:ECCV:2012}} \\ \hline
AEE     & 0.1191   & 0.1713   & 1.3895    & 1.5138     & 2.9714   & -  & {\bf
1.0735}   \\ \hline
MSE     & 0.3645   & 0.5979   & 6.2693    & 7.6105     & 21.4982  & -  & {\bf
3.2342}   \\ \hline
FE (\%) & 0.8155   & 1.1922   & 22.6290   & 21.9625    & 49.0136  & -  & {\bf
14.9061}   \\ \hline
\end{tabular}
\end{adjustbox}
\end{table*}
%

\begin{figure*}[ht]
\begin{center}
\begin{tabular}{cccc}
\includegraphics[width=0.2220\textwidth]{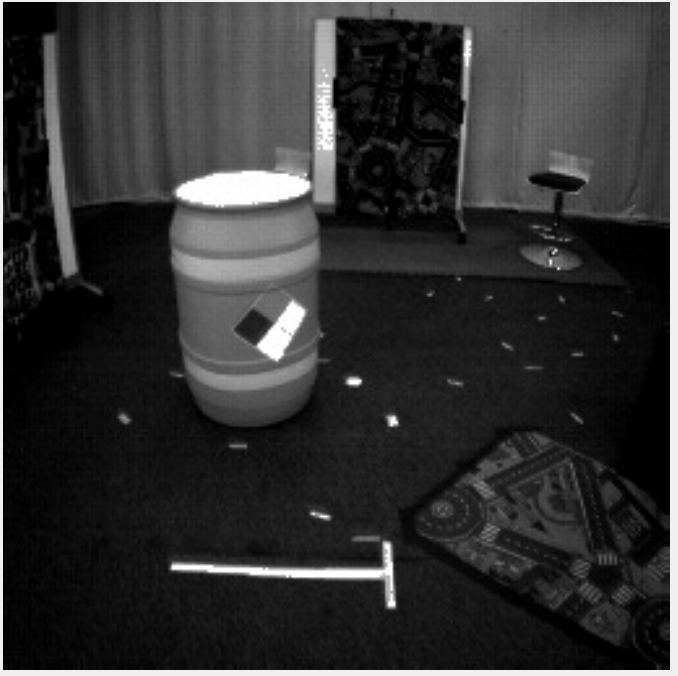}
&\includegraphics[width=0.2219\textwidth,height=0.2219
\textwidth]{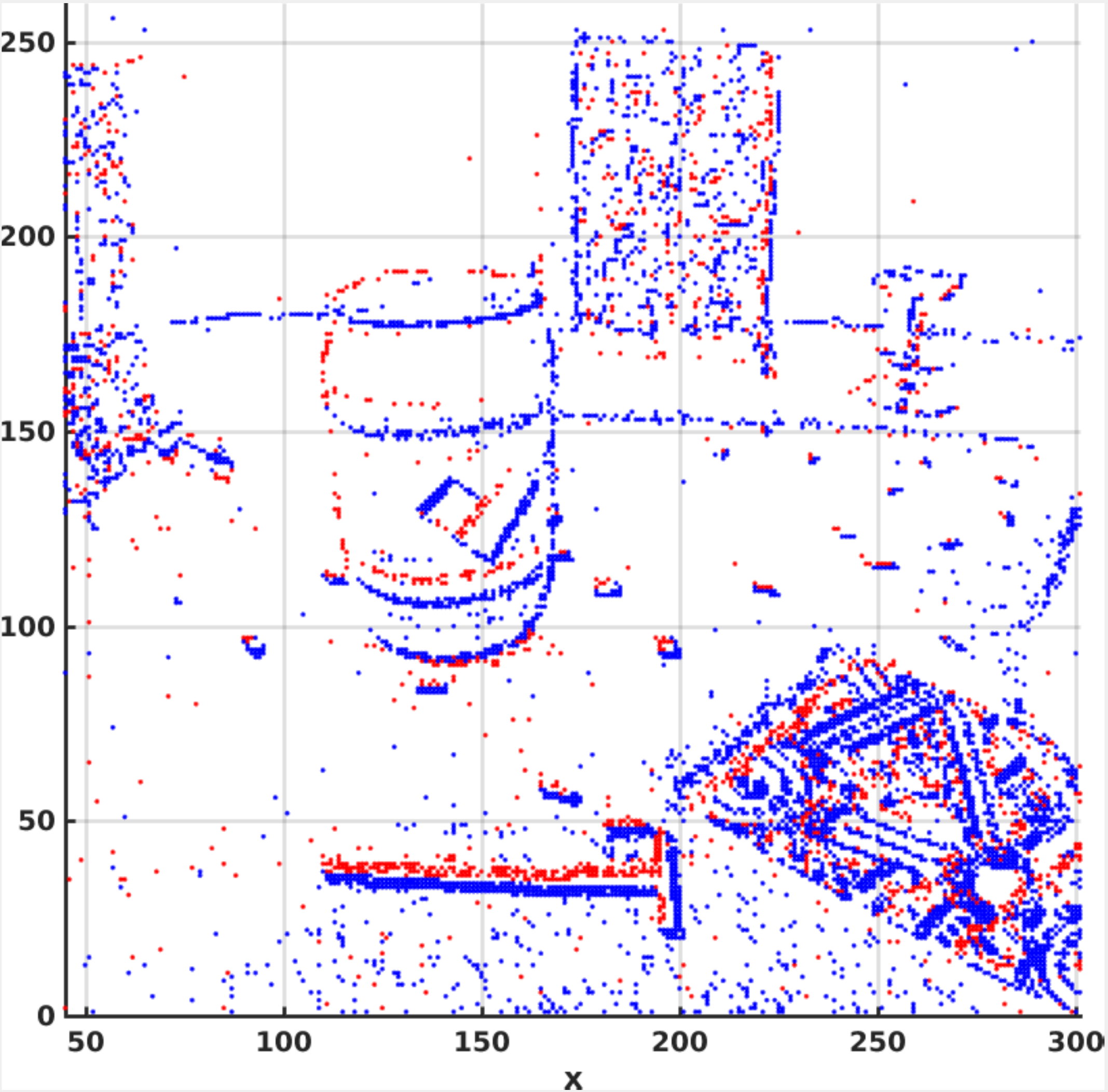}
&{\includegraphics[width=0.2219\textwidth]{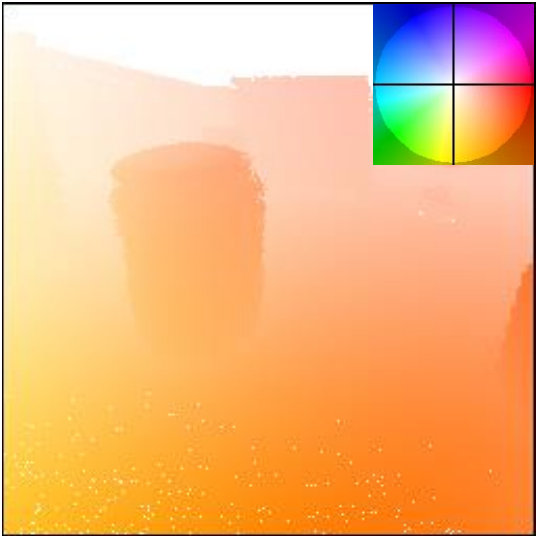}}
&{\includegraphics[width=0.2219\textwidth,height=0.2219
\textwidth]{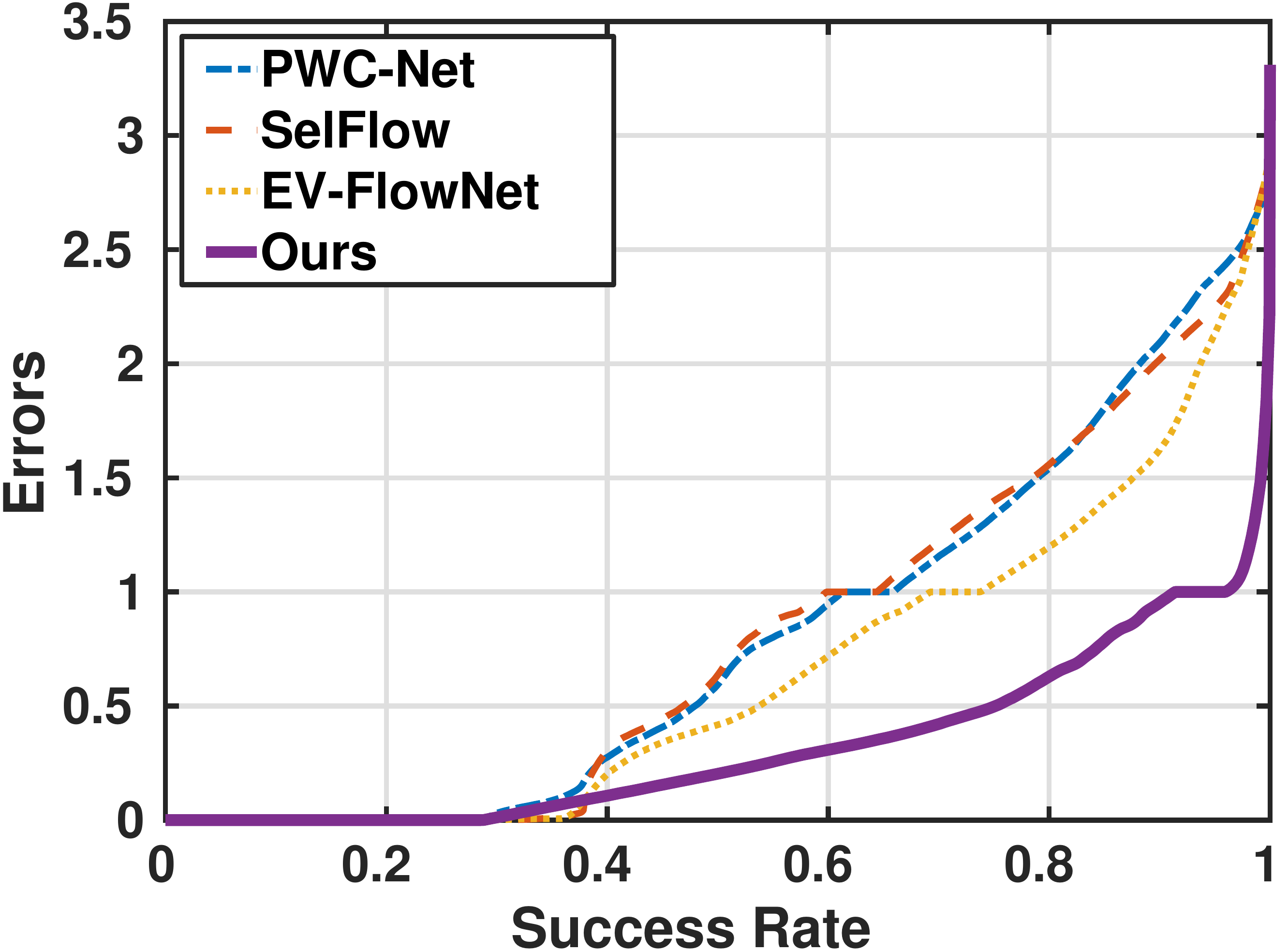}}\\
(a) Input image
&(b) Input events
&(c) Ground-truth flow
&(d) Error map\\
\includegraphics[width=0.2219\textwidth]{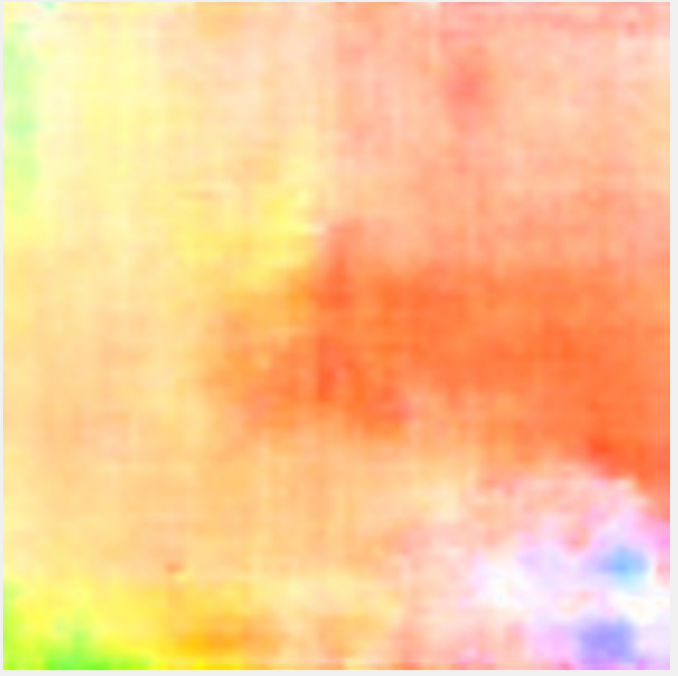}
&\includegraphics[width=0.2219\textwidth]{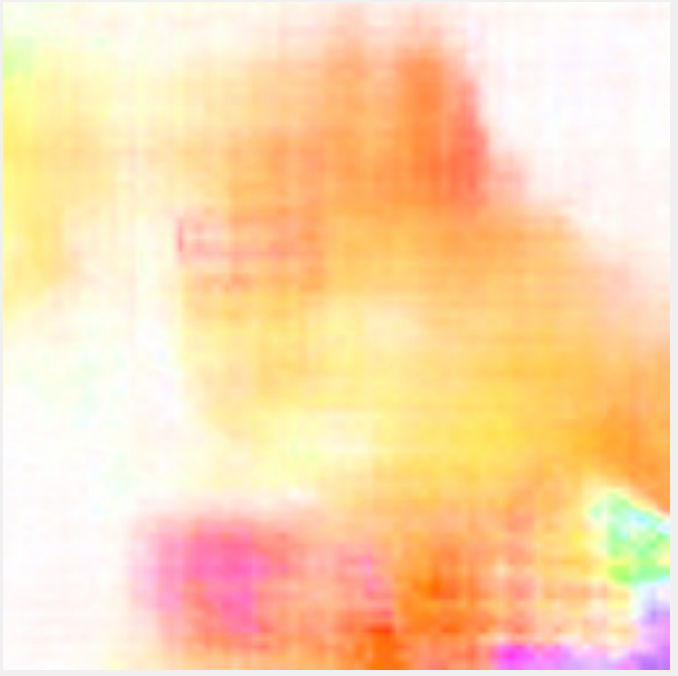}
&\adjincludegraphics[width=0.2219\textwidth,trim={0 {1.5mm} {1.5mm} 0},clip]{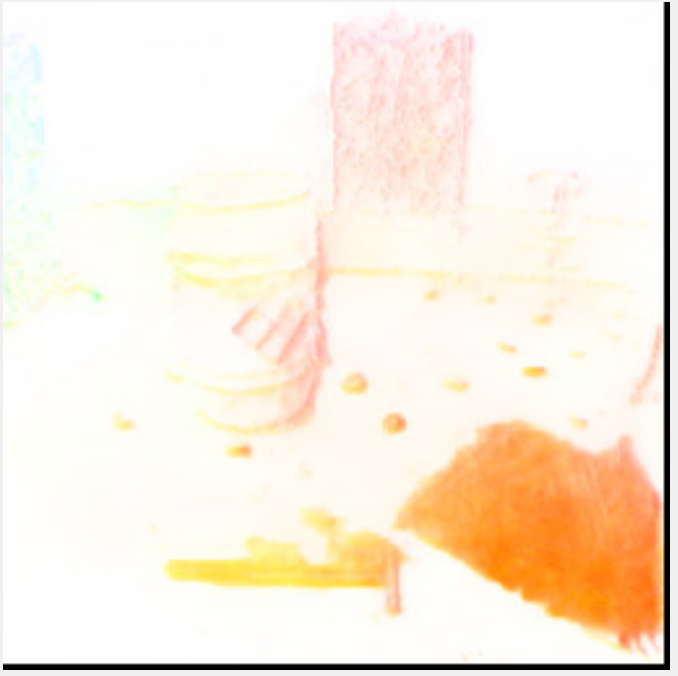}
&\includegraphics[width=0.2219\textwidth]{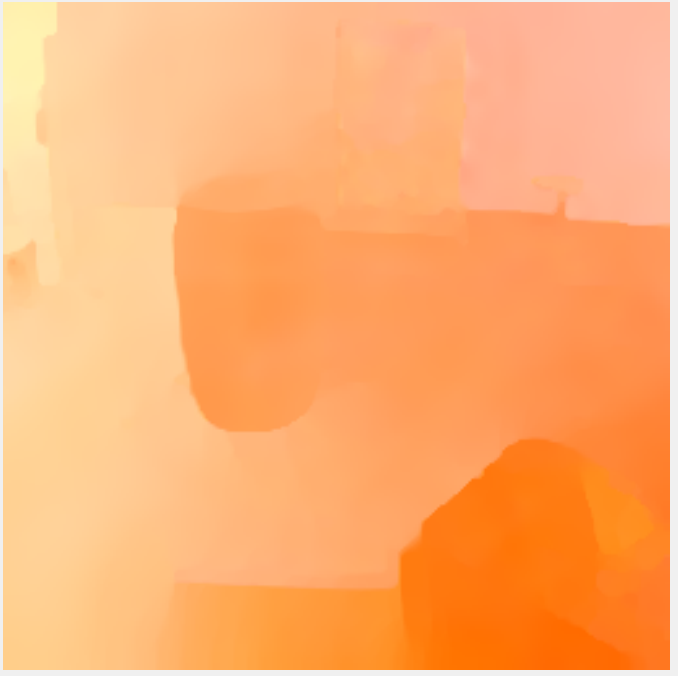}\\
(e) EDI + PWC-Net~\cite{Sun_2018_CVPR} 
&(f) EDI + SelFlow~\cite{Liu_2019_CVPR}
&(g) EV-FlowNet~\cite{Zhu-RSS-18}
&(h) Ours\\
\end{tabular}
\end{center}
\vspace*{-2 mm}
\caption{\label{fig:alex} \em Results of our method compared with
state-of-the-art methods on real dataset \cite{Zhu-RSS-18}. (a) Input image. (b)
Input events. (c) Ground-truth optical flow and the colour coded optical flow on
the left corner. (d) Error Map shows the distribution of the Endpoint Error of
estimates compared with the ground-truth flow. (e) Baseline: Flow result by
\cite{Sun_2018_CVPR} based on two reconstructed images. The reconstructed image
is estimated by EDI model \cite{Pan_2019_CVPR} from a single image and its
events. (f) Baseline: Flow result by \cite{Liu_2019_CVPR} based on two
reconstructed images. (g) Flow result by \cite{Zhu-RSS-18} based on images and
events. (h) Ours, by using an image and events as input. (Best viewed on
screen).}
\end{figure*}
\begin{table}[ht]\footnotesize
\centering
\caption{ \label{tab:Ablation}
\em Ablation Study based on Sintel Dataset \cite{Butler:ECCV:2012}.
}
\begin{tabular}{C{1.5cm}|C{2.5cm}|C{2.5cm}}
\hline
          & without $\phi_\textrm{eve}$    & without $\phi_\textrm{blur}$    
\\ \hline
AEE     & 2.3941   & 2.2594\\ \hline
MSE     & 5.3506   & 9.5267\\ \hline
FE (\%) & 18.0525  & 45.4516\\ \hline
\end{tabular}
\end{table}
%
%
\begin{table*}
\caption{ \label{tab:GoPro}
\em Quantitative analysis on the GoPro dataset \cite{Nah_2017_CVPR}. This
dataset provides ground-truth latent images and the associated motion blurred
images. The ground-truth optical flow is estimated by PWC-Net from the sharp video.
To demonstrate the efficiency of our optimization method, we use the output of 
`EDI + PWC-Net' as the input to our method. Our optimization method can still
show improvements. 
}
\centering
\begin{adjustbox}{width=0.98\textwidth}
\small
\begin{tabular}{c|c|c|c|c|c|C{2cm}}
\hline
Input     & \multicolumn{2}{c|}{EDI images and
events} & Events & \multicolumn{3}{c}{Image and
events}    \\ \hline
     & SelFlow~\cite{Liu_2019_CVPR}   &  PWC-Net \cite{Sun_2018_CVPR}  &
EV-FlowNet~\cite{Zhu-RSS-18}
     & EDI + PWC-Net + Our optimization
     & Our initialization & Our results             \\ \hline
AEE     & 2.0557  & 1.5806   & 2.0337   & 0.9796   & 3.7868    & {\bf 0.8641}   
 \\ \hline
MSE     & 5.7199  & 4.8951   & 10.5480  & 2.5952   & 8.3929    & {\bf 2.1536}   
 \\ \hline
FE(\%)  & 0.1722  & 0.1049   & 0.2839   & 0.0895   & 0.1218    & {\bf 0.0632}   
\\ \hline 
PSNR    & -       & -        & -        & 31.5595  & 29.3789  & {\bf 31.9234}  
\\ \hline
\end{tabular}
\end{adjustbox}
\end{table*}
%
\begin{figure*}[ht]
\begin{center}
\begin{tabular}{ccccc}
\hspace{-2.2 mm}
\includegraphics[width=0.1901\textwidth]{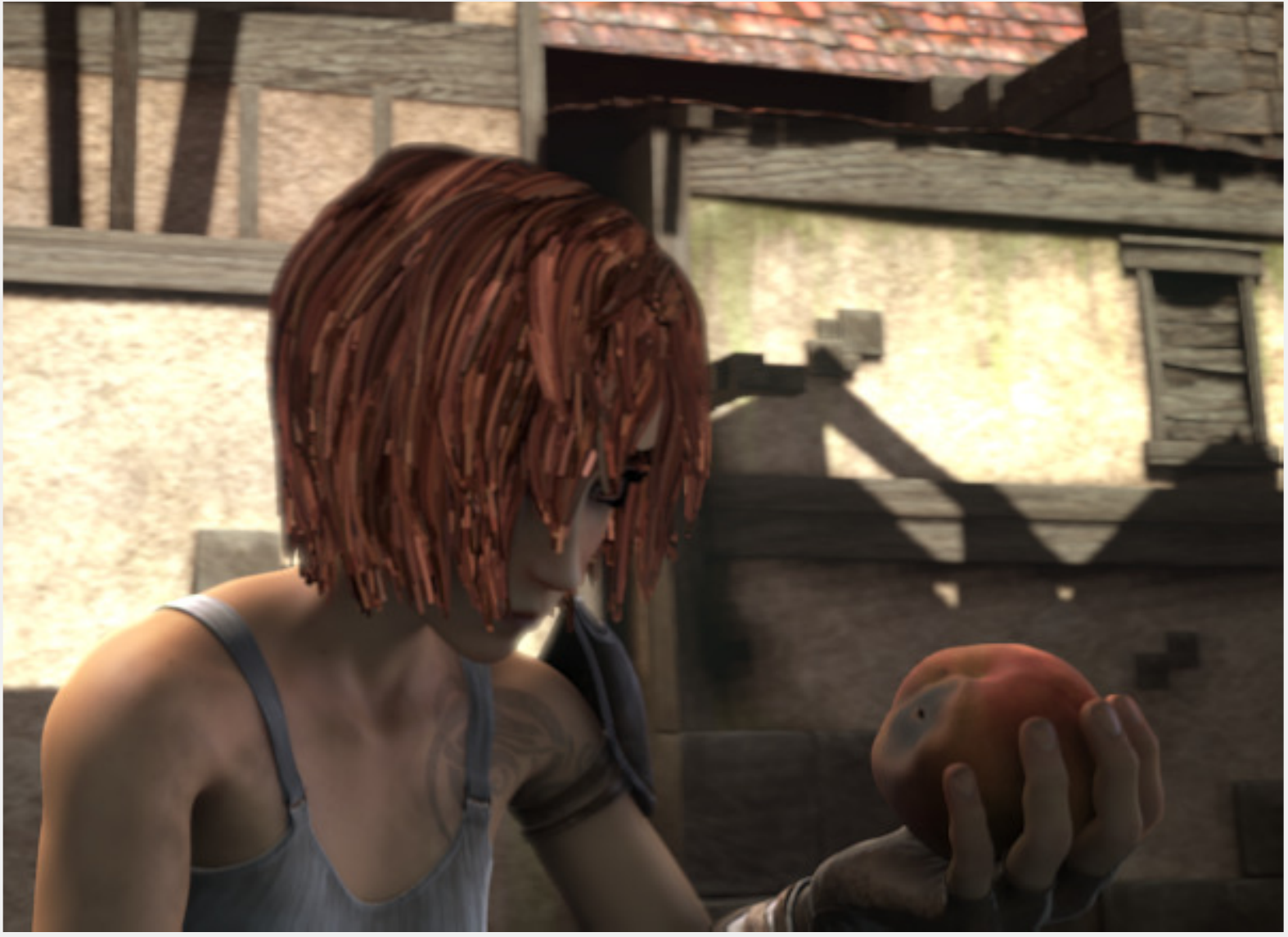}
\hspace{-3.5 mm}
&\includegraphics[width=0.1901\textwidth]{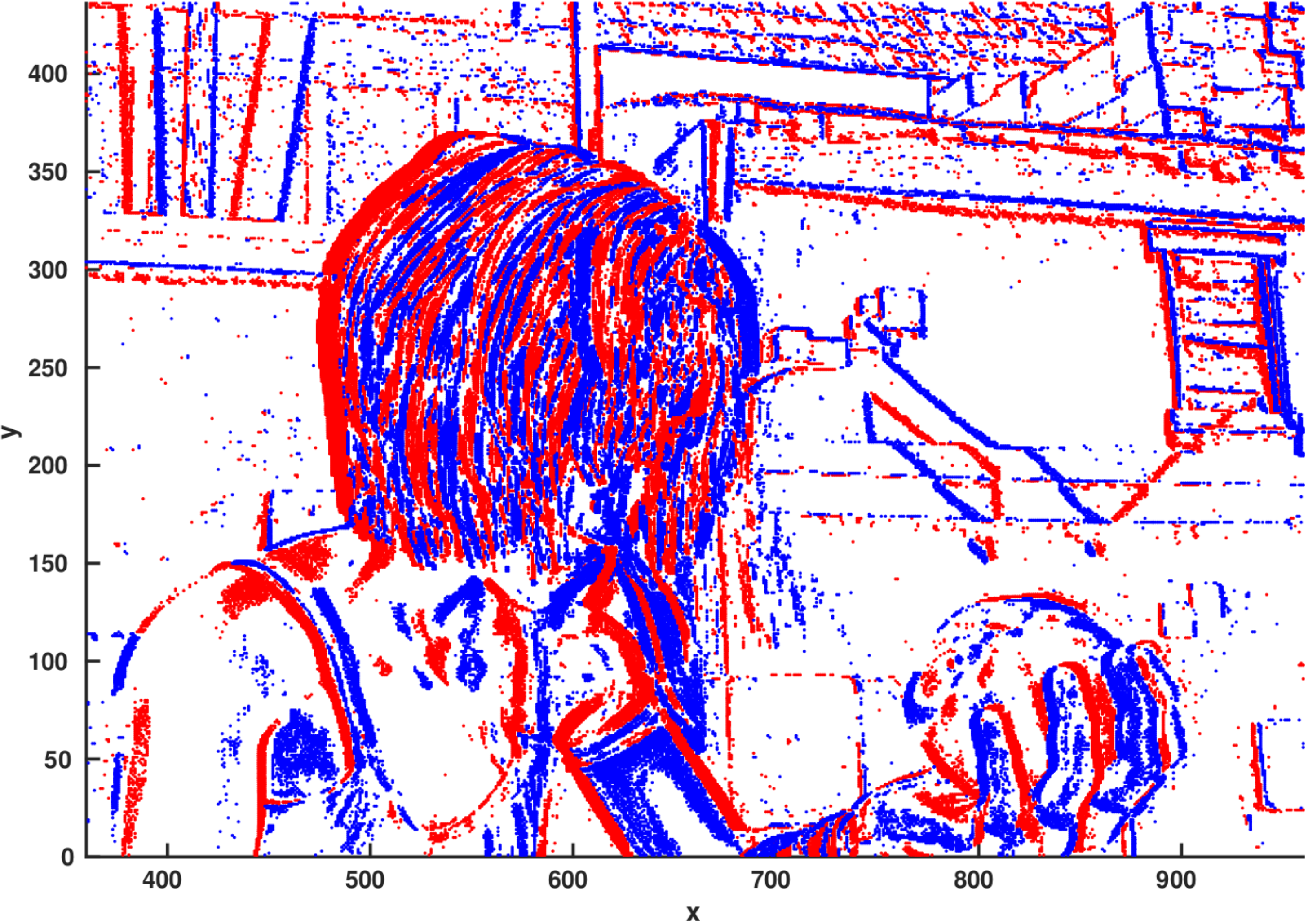}
\hspace{-3.5 mm}
&\includegraphics[width=0.1901\textwidth]{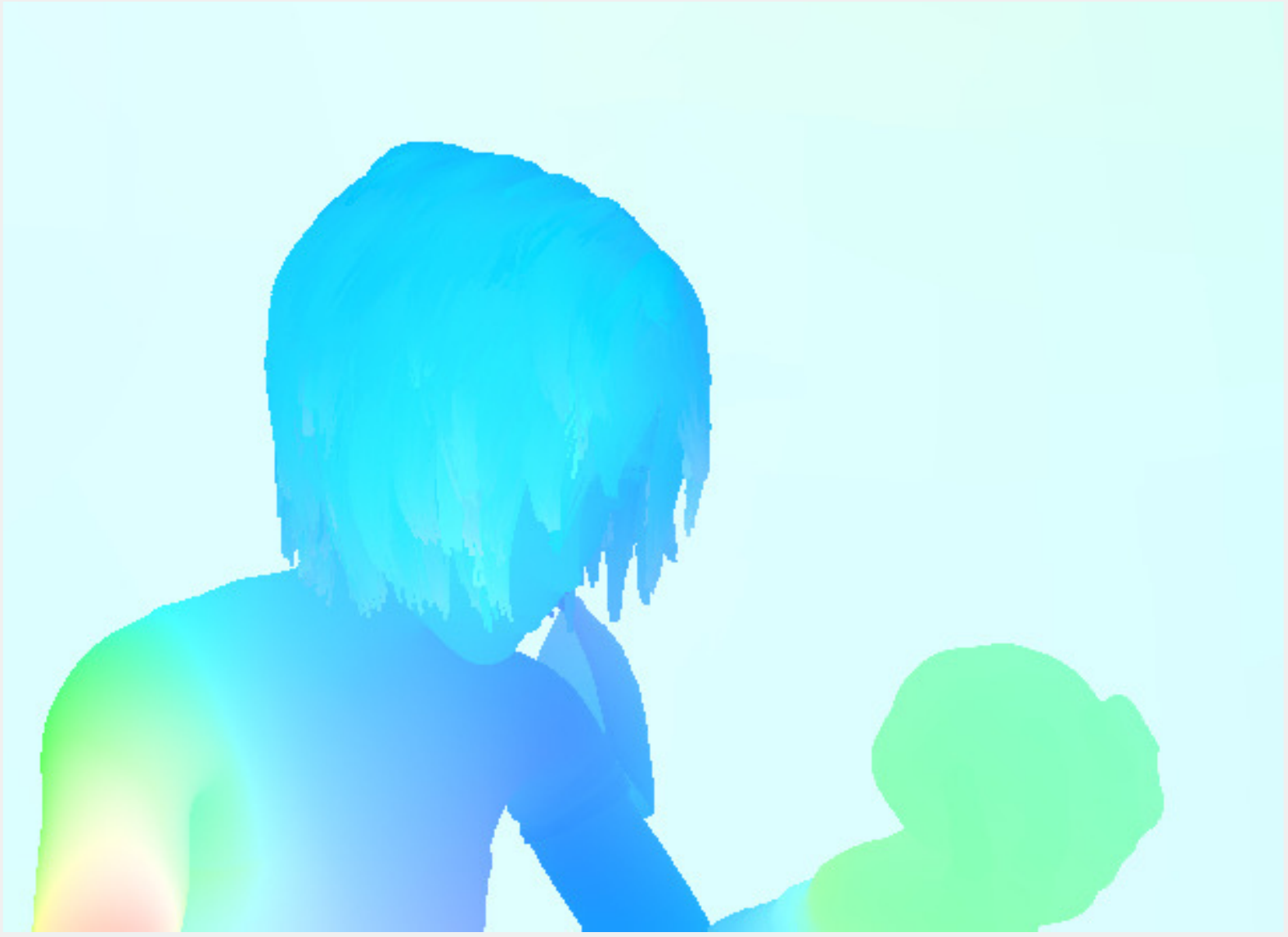}
\hspace{-3.5 mm}
&\includegraphics[width=0.1901\textwidth]{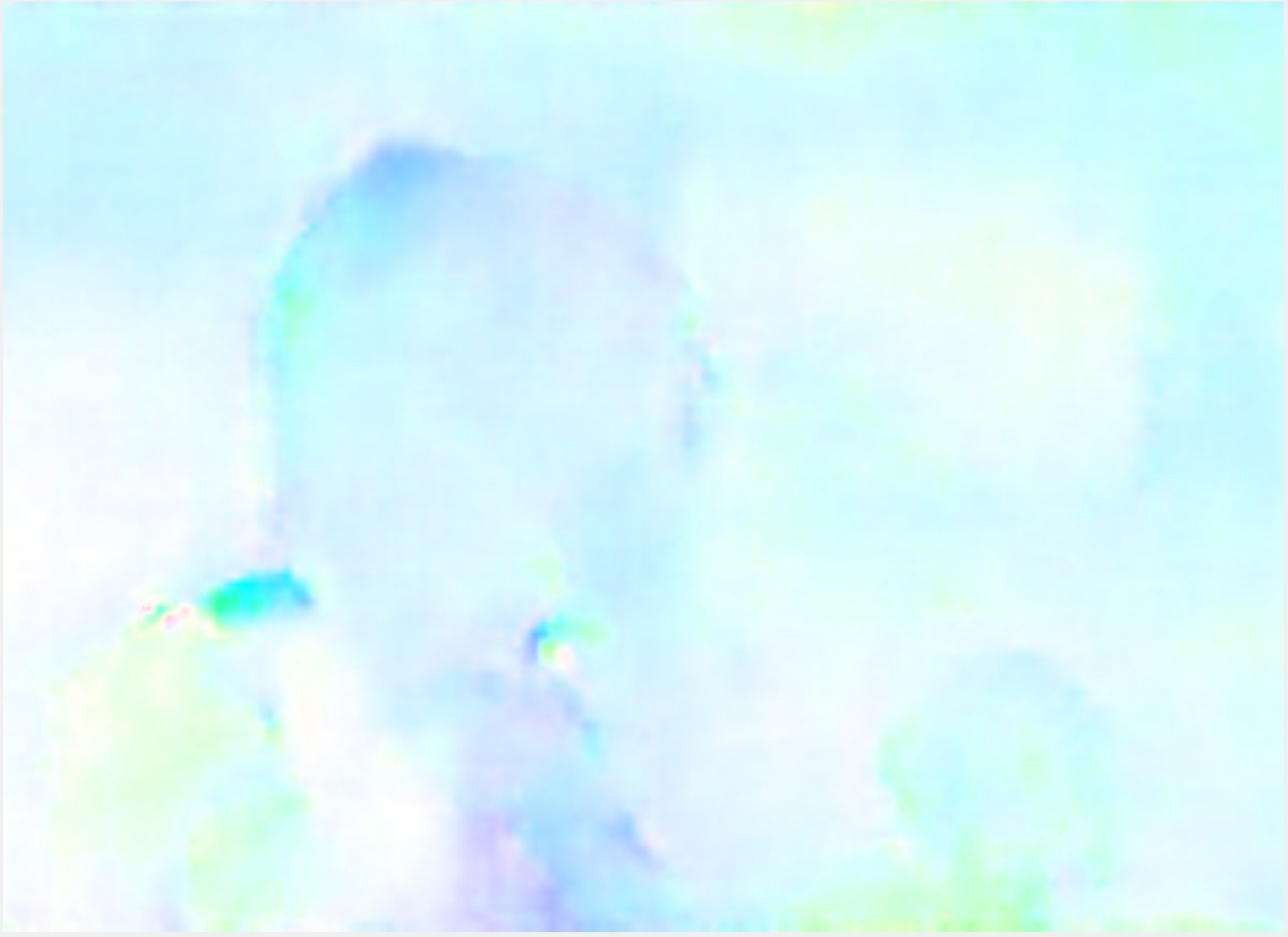}
\hspace{-3.5 mm}
&\includegraphics[width=0.1901\textwidth]{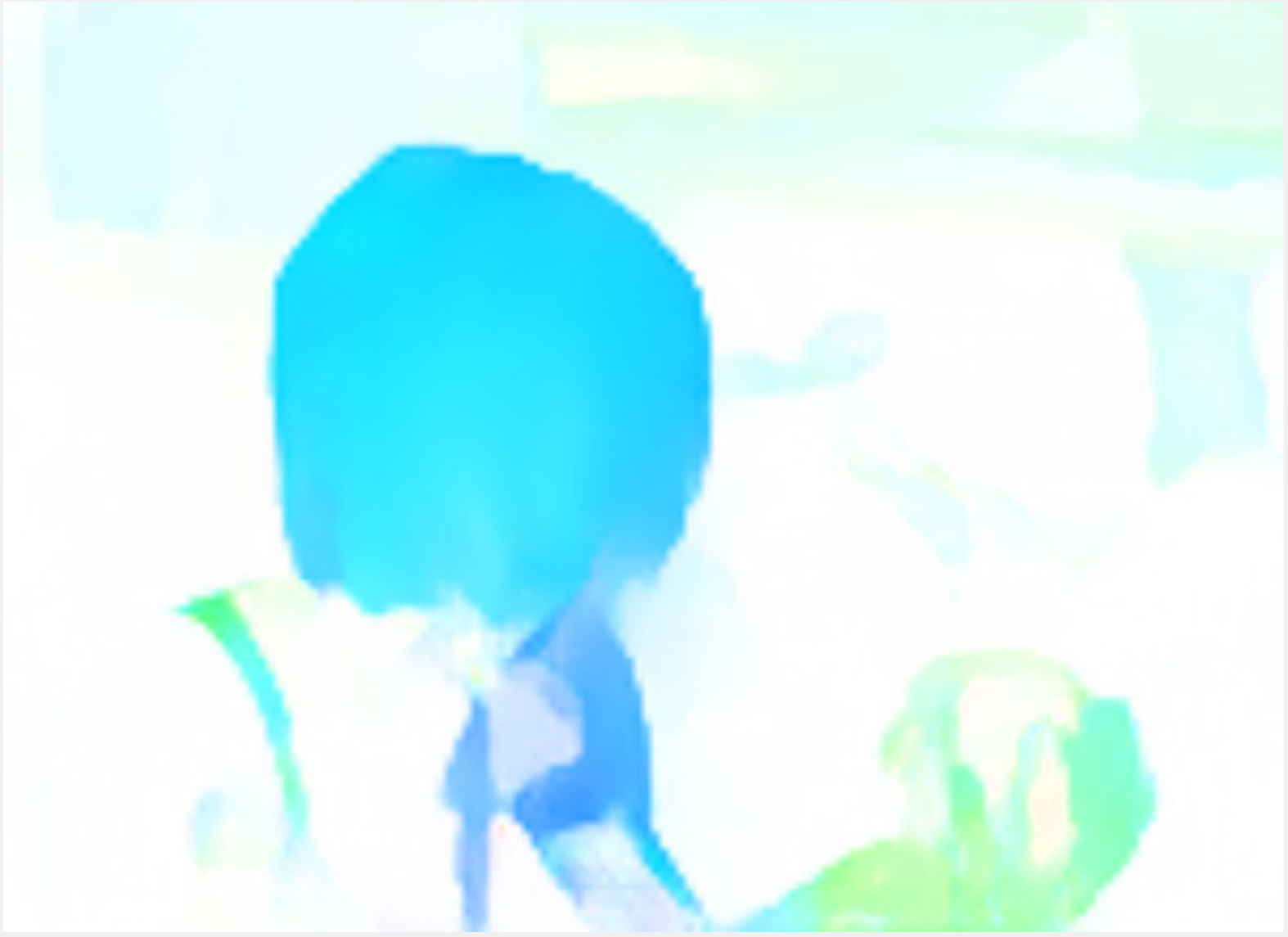}\\
\hspace{-2.2 mm}
(a) Input image 
\hspace{-3.5 mm}
&(b) Input events 
\hspace{-3.5 mm}
&(c) Ground-truth flow 
\hspace{-3.5 mm}
&(d) EDI + PWC-Net \cite{Sun_2018_CVPR}
\hspace{-3.5 mm}
&(e) EDI + SelFlow~\cite{Liu_2019_CVPR}\\
\hspace{-2.2 mm}
\includegraphics[width=0.1901\textwidth]{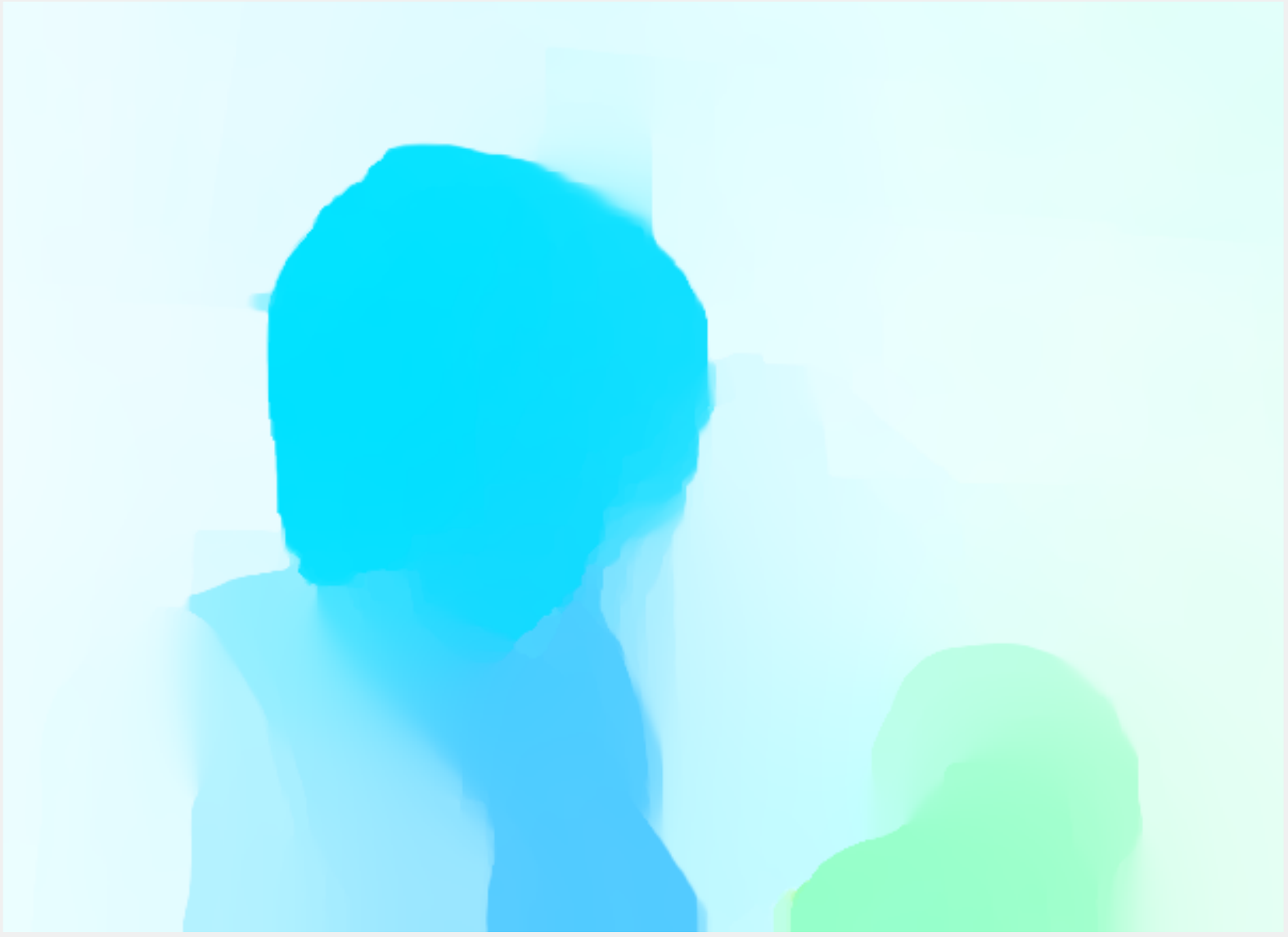}
\hspace{-3.5 mm}
&\includegraphics[width=0.1901\textwidth]{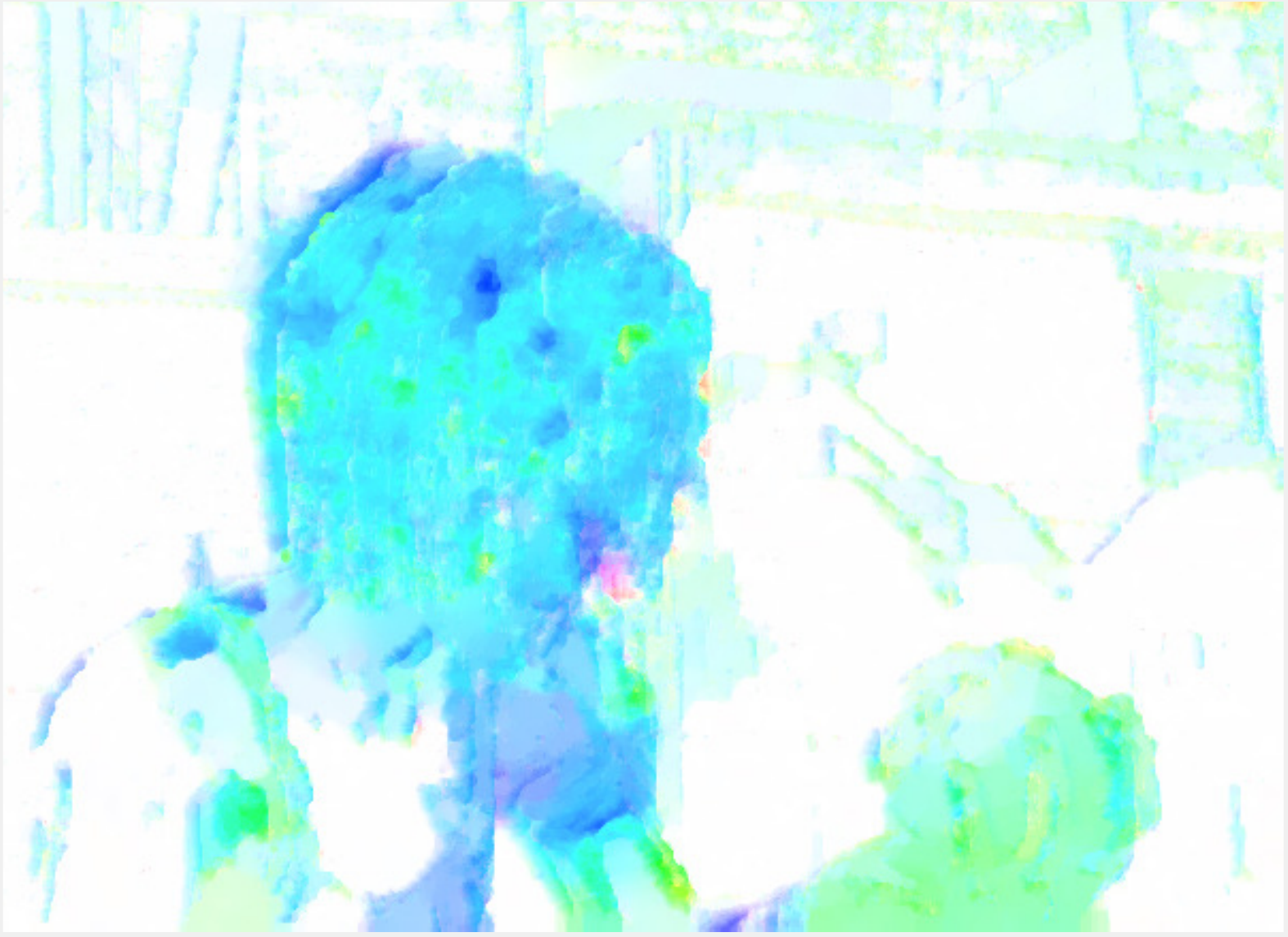}
\hspace{-3.5 mm}
&\includegraphics[width=0.1901\textwidth,height=0.142
\textwidth]{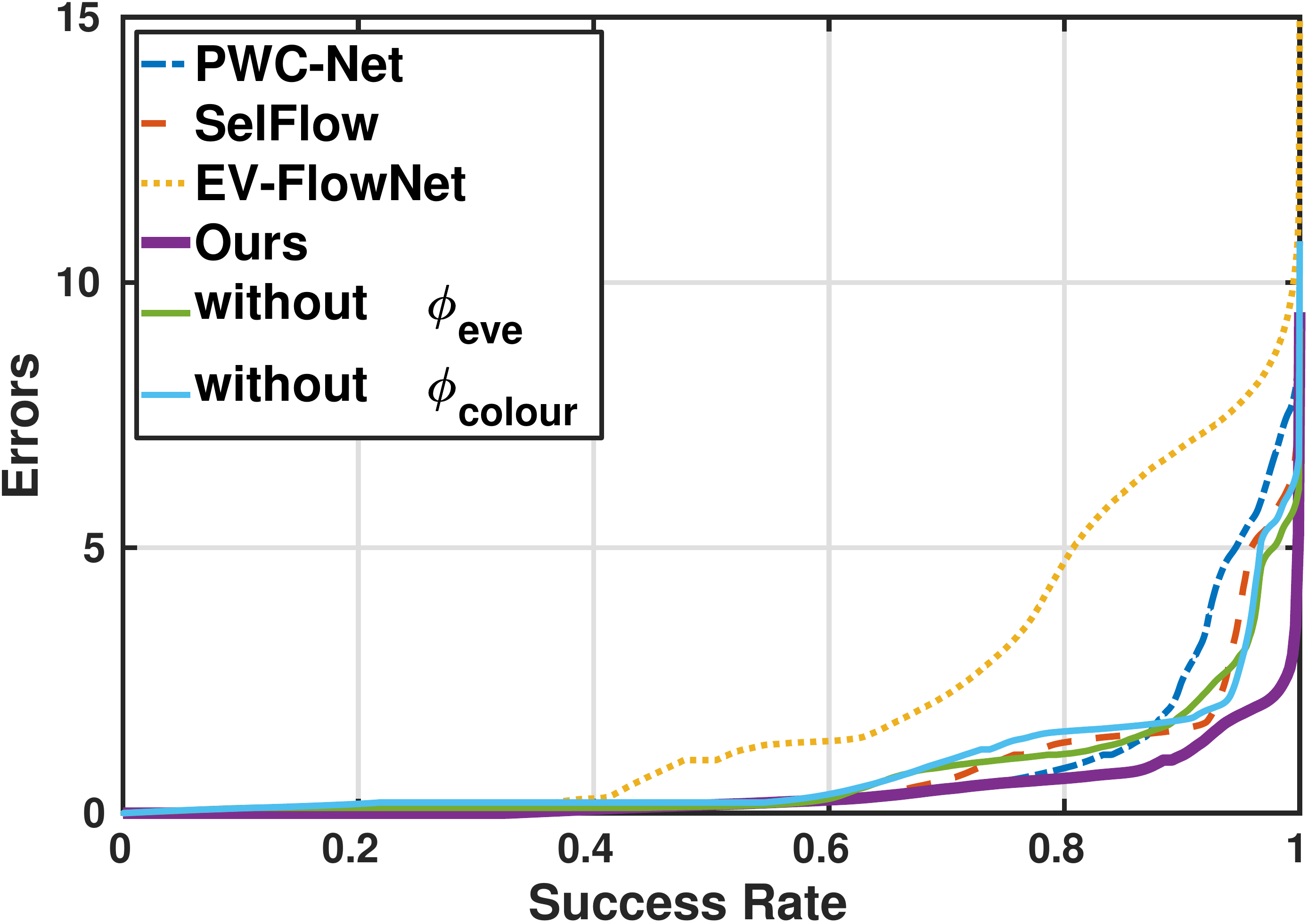}
\hspace{-3.5 mm}
&\includegraphics[width=0.1901\textwidth]{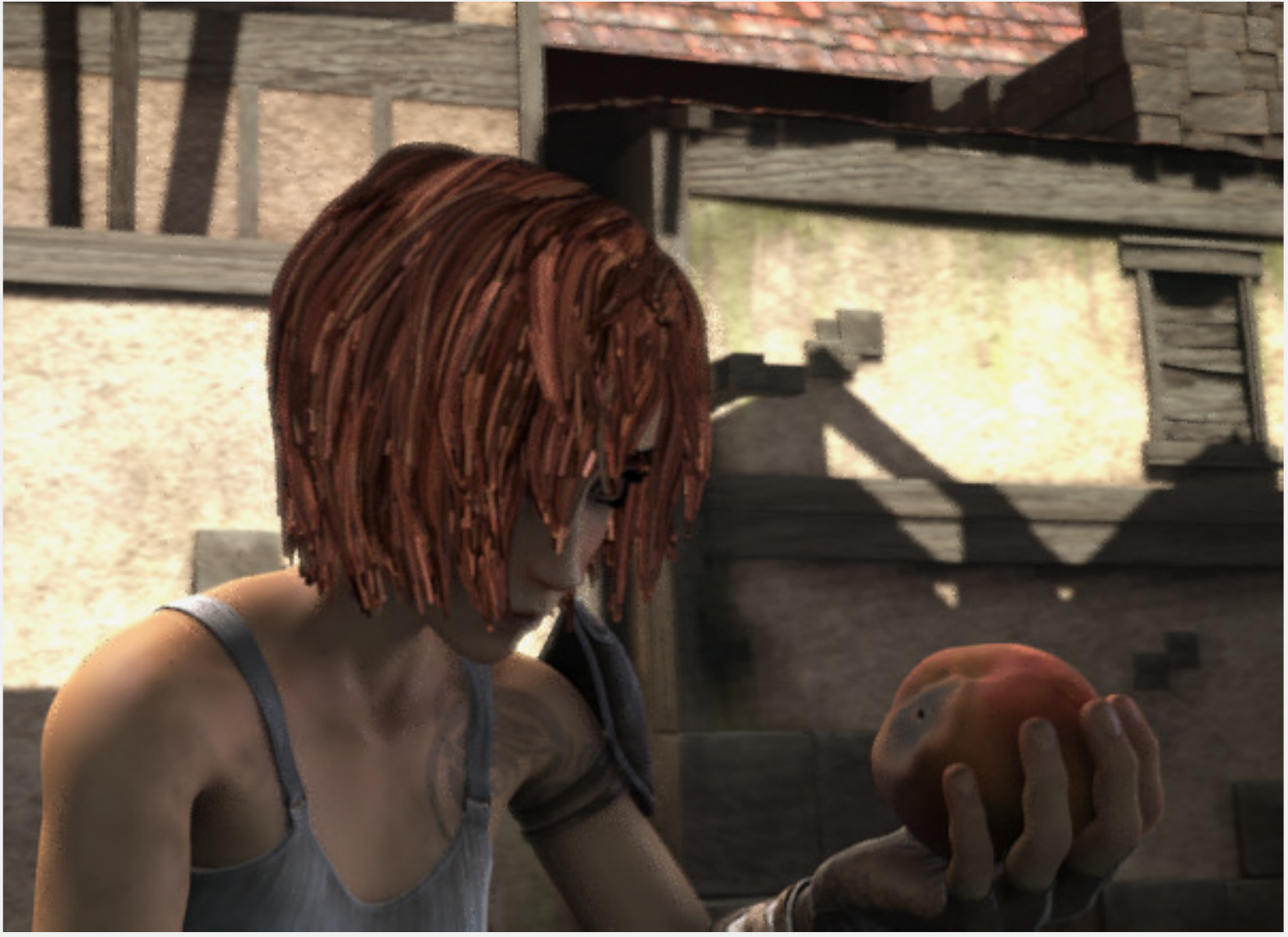}
\hspace{-3.5 mm}
&\includegraphics[width=0.1901\textwidth]{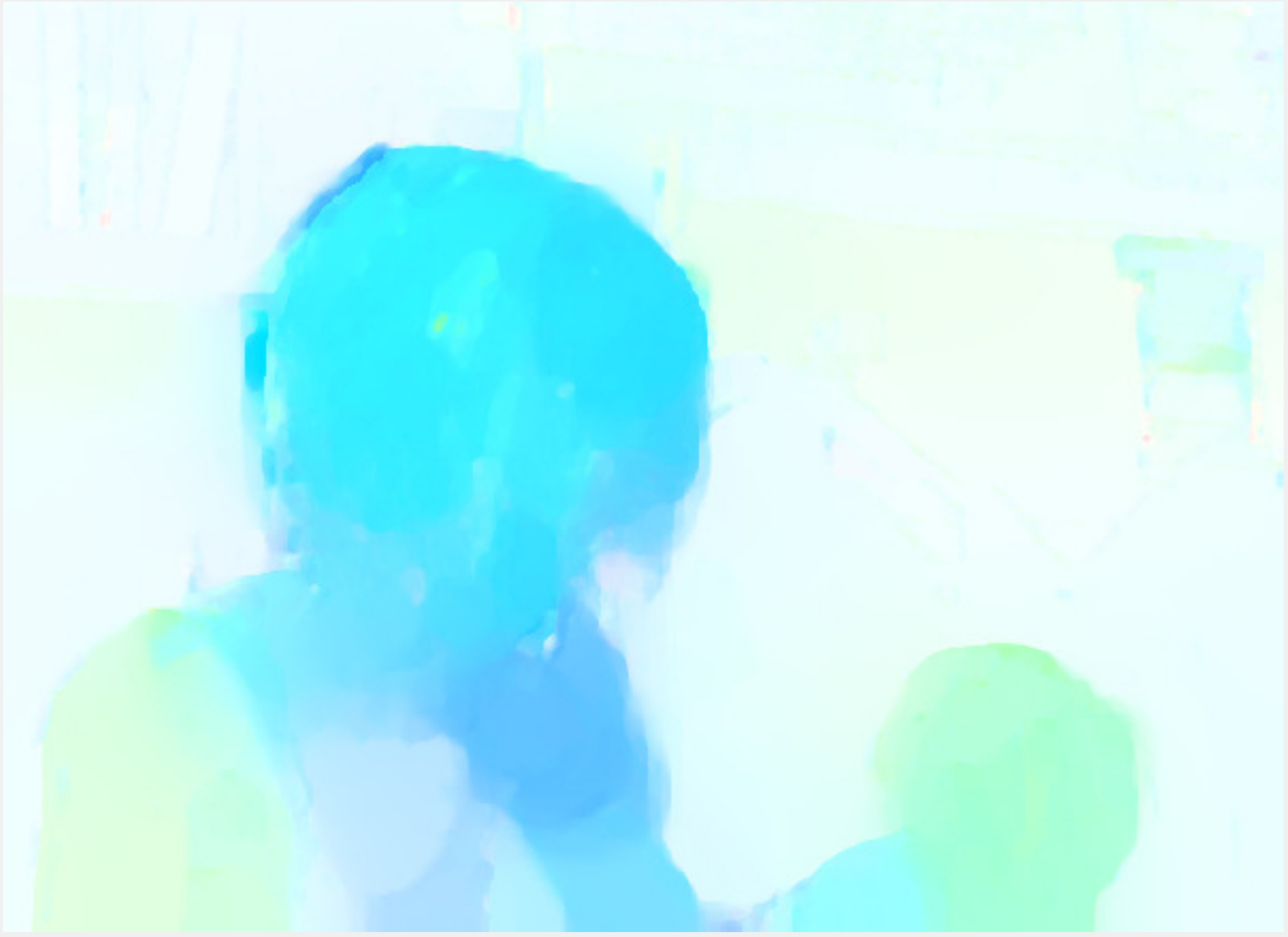}\\
\hspace{-2.2 mm}
(f) Ours - Without $\phi_\textrm{eve}$
\hspace{-3.5 mm}
&(g) Ours - Without $\phi_\textrm{blur}$
\hspace{-3.5 mm}
&(h) Error map
\hspace{-3.5 mm}
& (i) Ours - deblurred
\hspace{-3.5 mm}
& (j) Ours - optical flow \\
\end{tabular}
\end{center}
\vspace*{-2 mm}
\caption{\label{fig:flow0002sintel} \em An example of our method on dataset
\cite{Butler:ECCV:2012}. (a) Input blurred image. (b) Input events. (c)
Ground-truth optical flow. 
(d) Flow
result by \cite{Sun_2018_CVPR} based on images estimated by EDI model
\cite{Pan_2019_CVPR}. (e) Flow result by \cite{Liu_2019_CVPR} based on images
estimated by EDI model. (f) Ours baseline result without term
$\phi_\textrm{eve}$. (g) Ours baseline result without term
$\phi_\textrm{blur}$. (h) Error Map.  
(i) Our deblurring result. (j) Our optical flow. 
}
\end{figure*}
%
\begin{figure}
\begin{center}
\begin{tabular}{ccccc}
\hspace{-0.25 cm}
\includegraphics[width=0.0870\textwidth]{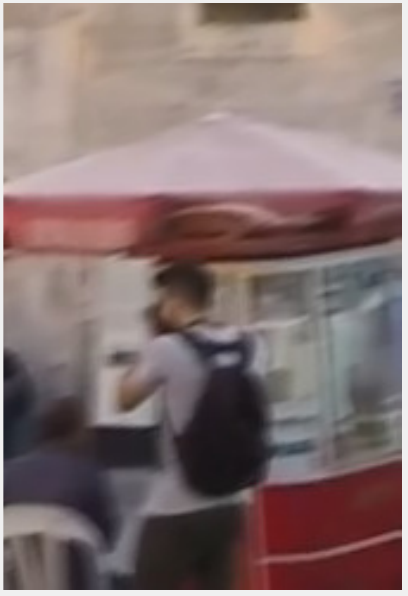}
\hspace{-0.35 cm}
&\includegraphics[width=0.0870\textwidth]{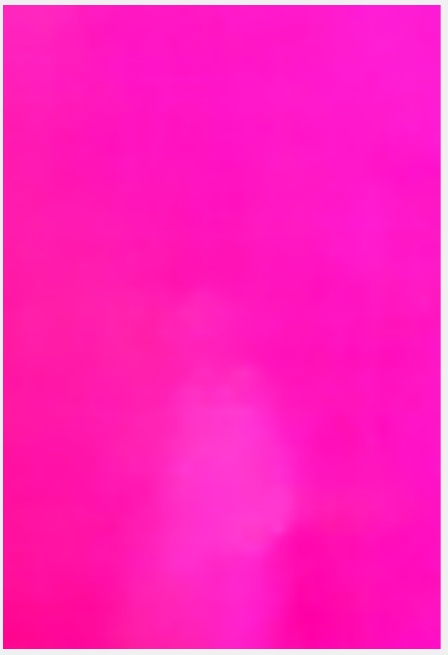}
\hspace{-0.35 cm}
&\includegraphics[width=0.0870\textwidth,height=0.1266\textwidth,trim={0 0 0 {1.5mm}},clip]{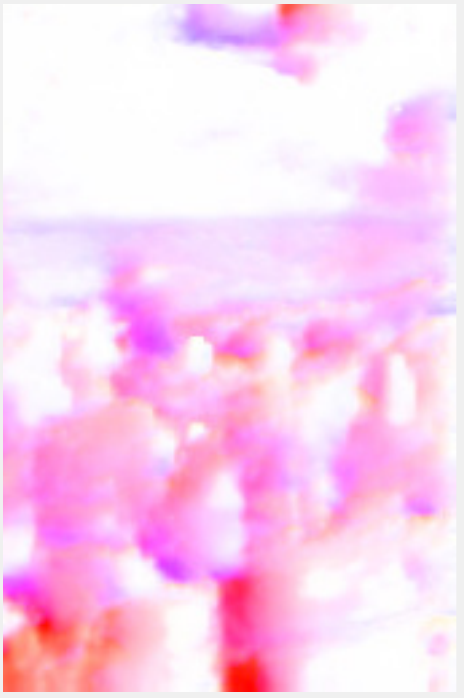}
\hspace{-0.35 cm}
& \includegraphics[width=0.0870\textwidth]{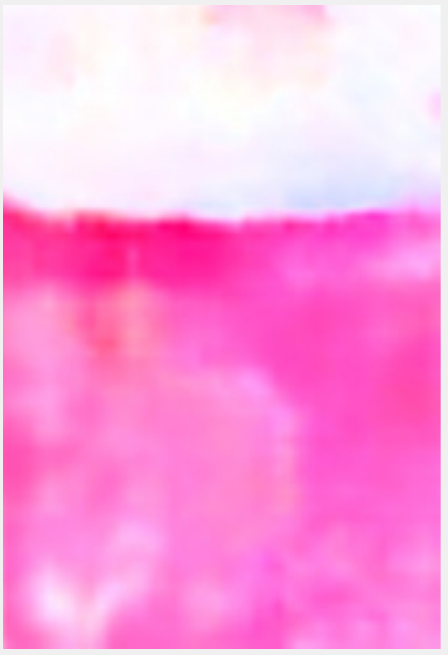}
\hspace{-0.35 cm}
& \includegraphics[width=0.0870\textwidth]{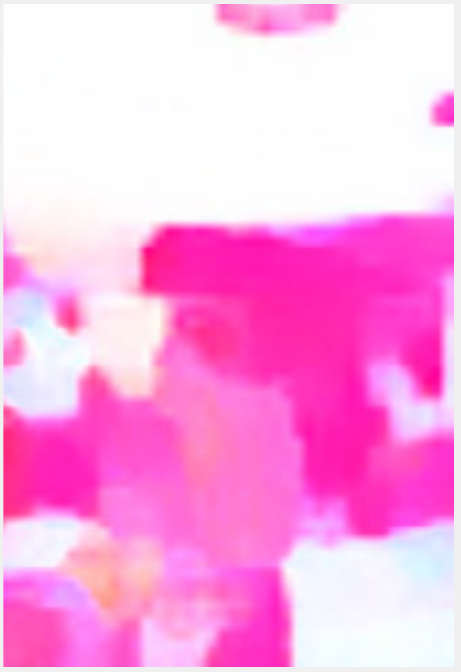}\\
\hspace{-0.25 cm}
(a) 
\hspace{-0.35 cm}
&(b) 
\hspace{-0.35 cm}
&(c) 
\hspace{-0.35 cm}
&(d) 
\hspace{-0.35 cm}
&(e) \\
\hspace{-0.25 cm}
\includegraphics[width=0.0870\textwidth]{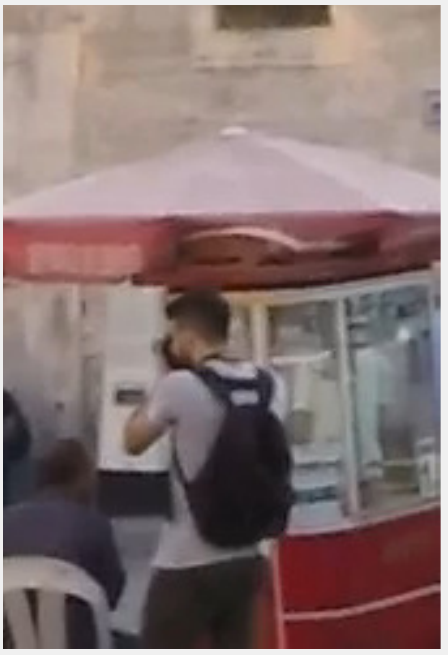}
\hspace{-0.35 cm}
&\includegraphics[width=0.0870\textwidth]{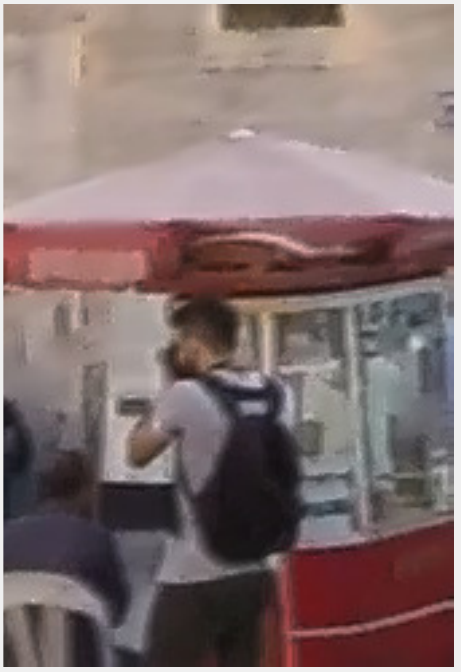}
\hspace{-0.35 cm}
&\includegraphics[width=0.0870\textwidth,height=0.1275\textwidth,trim={0 {1.5mm} 0 0},clip]{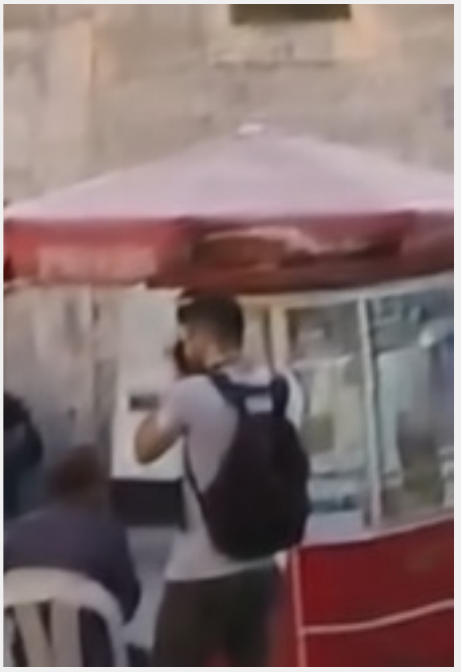}
\hspace{-0.35 cm}
& \includegraphics[width=0.0870\textwidth,height=0.1275\textwidth]{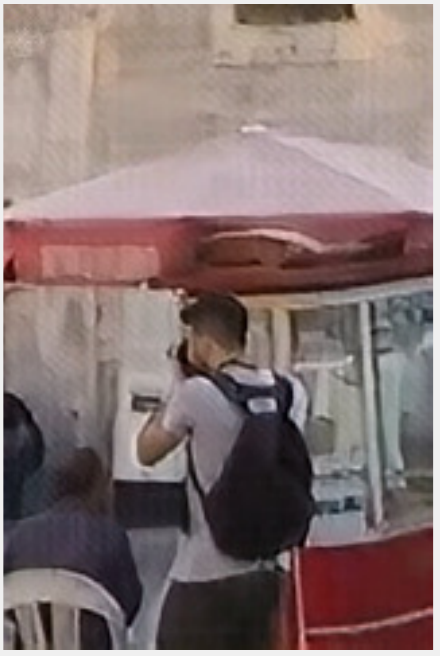}
\hspace{-0.35 cm}
& \includegraphics[width=0.0870\textwidth]{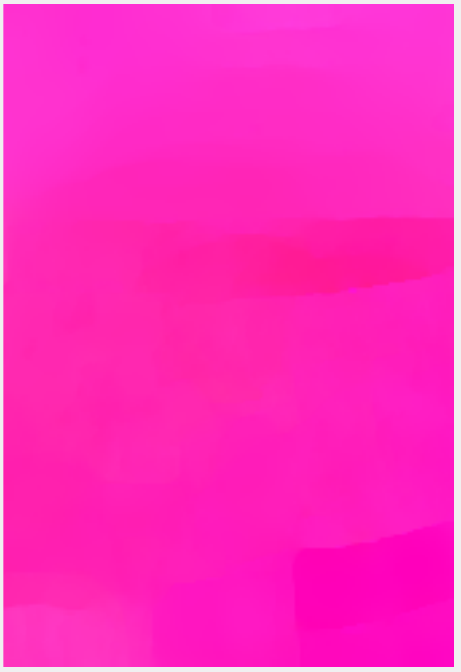}\\
\hspace{-0.25 cm}
 (f) 
\hspace{-0.35 cm}
&(g) 
\hspace{-0.35 cm}
&(h) 
\hspace{-0.35 cm}
&(i) 
\hspace{-0.35 cm}
&(j) \\
\end{tabular}
\end{center}
\vspace*{-2 mm}
\caption{\label{fig:nah} \em An example of our method on dataset
\cite{Nah_2017_CVPR}.
(a) The blurred image. 
(b) The ground-truth flow.
(c) Flow result by \cite{Zhu-RSS-18}, using the events as input. 
(d) Flow result by \cite{Sun_2018_CVPR} based on images estimated by the EDI model \cite{Pan_2019_CVPR}. 
(e) Flow result by \cite{Liu_2019_CVPR} based on images estimated by the EDI model. 
(f) The ground-truth latent images at time $t$.
(g)Deblurred result by \cite{Pan_2019_CVPR}. 
(h) Deblurred result by \cite{Zhang_2019_CVPR}.
(i) Our deblurred image. 
(j) Our estimated optical flow. 
}
\end{figure}

\subsection{Experimental Setup}
\noindent{\bf{Real dataset.}} 
We evaluate our method on three public real event datasets, namely, Multi-vehicle
Stereo Event Camera dataset (MVSEC)~\cite{zhu2018multivehicle}, Event-Camera
dataset (ECD)~\cite{mueggler2017event}, and Blurred Event Dataset
(BED)~\cite{Scheerlinck18accv,Pan_2019_CVPR}.
MVSEC provides a collection of sequences captured by DAVIS for high-speed
vehicles with ground truth optical flow. 

\vspace*{+1 mm}
\noindent{\bf{Synthetic dataset.}} 
For quantitative comparisons on optical flow, we build a synthetic dataset based
on Sintel~\cite{Butler:ECCV:2012} with images of size $1024 \times 436$,
which uses the event simulator ESIM~\cite{rebecq2018esim} to generate event
streams. While Sintel provides a blurred dataset, it mainly focuses on out of
focus blur instead of motion blur. Therefore, it is not suitable for the
evaluation of deblurring.~To provide a quantitative deblurring comparison, we
generate another synthetic dataset with events and motion blur, based on the
real GoPro video dataset \cite{Nah_2017_CVPR}, where the image size is $1280
\times 720$. It has ground-truth latent images and associated motion blurred
images.
We additionally use PWC-Net to estimate flow from sharp images as the
ground-truth for flow evaluation.

\vspace*{+1 mm}
\noindent{\bf Evaluations.}
For the evaluation of flow estimation results, we use error metrics, such as
Mean Square Error (MSE), Average Endpoint Error (AEE), and Flow Error metric (FE)
(seeing details in the supplementary material). FE metric is computed by
counting the number of pixels having errors more than $3$ pixels and $5\%$ of
its ground-truth over pixels with valid ground truth flow. We adopt the PSNR to
evaluate deblurred images. The error map shows the distribution of the endpoint
error of measurements compared with the ground-truth flow and the success rate
is defined as the percentage of results with errors below a threshold.

\vspace*{+1 mm}
\noindent{\bf Baseline methods.}
For optical flow, we compare with state-of-the-art event only based methods EV-FlowNet~\cite{Zhu-RSS-18}, and Zhu \etal \cite{zhu2019unsupervised}.
Then, we compare with the state-of-the-art video (with the label `GT images'.) only based method SelFlow~\cite{Liu_2019_CVPR}, and PWC-Net~\cite{Sun_2018_CVPR}. 
In addition, we build a two-step ( event + image) framework as a baseline approach, which is 'EDI + SelFlow' and 'EDI +  PWC-Net'. The two-step framework first use the image reconstruction method EDI~\cite{Pan_2019_CVPR} to restore intensity images, then applying flow estimation methods~\cite{Sun_2018_CVPR, Liu_2019_CVPR} to the restored images to estimate flow.
We compare our deblurring results with the state-of-the-art event-based deblurring approach~\cite{Pan_2019_CVPR} and blind deblurring methods \cite{Zhang_2019_CVPR,Tao_2018_CVPR,gong2017motion}. 

\vspace*{+1 mm}
\noindent{\bf{Implementation details.}} 
For all our real experiments, the image and events are from DAVIS. 
The framework is implemented by using MATLAB with C++ wrappers. It takes around
20 seconds to process a real image (size $346\times 260$) from DAVIS on a single
i7 core running at 3.6 GHz. 

\subsection{Experimental Results}
 
We compare our results with baselines on optical flow estimation and image
deblurring on 5 (including real and synthetic) datasets. Our goal is to demonstrate that given a single blurred image and event stream, jointly optimising the image and optical flow would achieve better results than ``event only'', ``single (blurred) image only'', and stage-wise methods.
We report quantitative comparisons in Table \ref{tab:MVSEC}, \ref{tab:GoPro} and
qualitative comparisons in Fig. \ref{fig:nightrun}, \ref{fig:eth},
\ref{fig:alex}, \ref{fig:flow0002sintel} to show the effectiveness and
generalization of our method. 
Ablation study in Table \ref{tab:Ablation} shows the effectiveness of each term
in our objective function \eqref{eq:energyfunc}. 

As shown in Table~\ref{tab:MVSEC} and Fig.~\ref{fig:alex}, we achieve
competitive results on flow estimation compared with event only based methods
\cite{Zhu-RSS-18,zhu2019unsupervised} on MVSEC dataset. 
Note that models in~\cite{Zhu-RSS-18,zhu2019unsupervised} are trained on MVSEC
while our model can still achieve competitive results without training. 
As BED and ECD do not provide ground-truth flow or sharp image for evaluation,
we thus show qualitative comparisons in Fig. \ref{fig:nightrun} and
\ref{fig:eth}, which demonstrate the stability of our model under both blurred
and non-blurred conditions.

We show flow comparisons in Table \ref{tab:MVSEC} and Fig.
\ref{fig:flow0002sintel} on the Sintel dataset. 
While Sintel provides a blurred dataset mainly focusing on
out-of-focus blur (including slightly motion blur), our method can achieve
competitive results on flow estimation. Also, we gained a 1 dB increase on the
PSNR metric for image deblurring. In Table~\ref{tab:GoPro} and Fig. \ref{fig:nah}, we provide deblurring comparisons on GoPro dataset \cite{Nah_2017_CVPR}. 
Our
approach outperform all the baseline methods on flow estimation and image
deblurring, 
which further indicated that 1) including a single image helps achieve better flow estimate than event only based approaches especially in regions with no events, 2) two-stages approaches suffer from image artifacts (even images from EDI) which motivate us to jointly perform image refinement and flow estimate.
More results can be found in the supplementary material. 

\vspace*{+1 mm}
\noindent{\bf Ablation Study.}
To provide a deep understanding of our model, we evaluate the influence of
$\phi_\textrm{eve}$ and $\phi_\textrm{blur}$ in Table \ref{tab:Ablation}. The
significantly decreased performance indicates the contribution of each term in
our model. In Table~\ref{tab:GoPro}, we add a comparison to demonstrate the
efficiency of our optimization strategy. With a better flow input from `EDI +
PWC-Net', we can still achieve significant improvement. Note, the threshold $c$ is estimated based on \cite{Pan_2019_CVPR} and our initial flow
is simply computed using Eq.~\eqref{eq:horn} on event frames.

\section{Conclusion}
In this paper, we jointly estimate optical flow and the sharp intensity image based on a single image (potentially blurred) and events from DAVIS. Under our formulation, events are high-efficiency data that can reinforce flow estimation. Extensive experiments on different datasets produce competitive results that show the generalization ability, effectiveness and accuracy of our model. While our approach can handle high dynamic cases, we still have difficulties in tackling low texture scenarios, and unstably with noise event data like other methods. Our future work will explore events representation to build a learning-based end-to-end flow estimation Neural Network with the image.

\section{Acknowledgment}
This research was supported in part by the Australian Research Council through the “Australian Centre of Excellence for Robotic Vision” CE140100016, and the (ARC) fellowship and Discovery Project grant (DE180100628, DP200102274).

{\small
\bibliographystyle{ieee}
\bibliography{cvpr20-bib-eventwithdeep}
}

\end{document}